\documentclass[twocolumn,switch,times]{article}
\usepackage{preprint}
\usepackage{hyperref}
\usepackage[authoryear,round]{natbib}

\hypersetup{
    colorlinks=true,
    linkcolor=blue,
    urlcolor=blue,
    citecolor=blue,
    anchorcolor=black
}

\usepackage[utf8]{inputenc}	
\usepackage[T1]{fontenc}
\usepackage{xcolor}
\usepackage{lineno}					
\usepackage{tikz} 					
\usepackage{fancyhdr}
\usepackage{geometry}
\usepackage{lipsum}
\usepackage{scalerel}
\usepackage{sidecap}
\usepackage{times}
\usepackage{doi}
\usepackage{amssymb}
\usepackage{amsmath}
\usepackage{bbm}
\usepackage{bm}
\usepackage{array,booktabs}
\usepackage{upgreek}
\usepackage{algorithm}
\usepackage{algpseudocodex}
\usepackage{appendix}
\usepackage{authblk}
\usepackage{placeins}

\usepackage{newfloat}
\DeclareFloatingEnvironment[name={Supplementary Figure}]{suppfigure}
\usepackage{sidecap}
\sidecaptionvpos{figure}{c}
\usepackage{etoolbox}

\usepackage{titlesec}
\titlespacing\section{0pt}{12pt plus 3pt minus 3pt}{1pt plus 1pt minus 1pt}
\titlespacing\subsection{0pt}{10pt plus 3pt minus 3pt}{1pt plus 1pt minus 1pt}
\titlespacing\subsubsection{0pt}{8pt plus 3pt minus 3pt}{1pt plus 1pt minus 1pt}

\definecolor{lime}{HTML}{A6CE39}

\newcommand{\indicatorpred}[2]{\mathbbm{1}(\hat{#1}_{#2}=#1_{#2})}
\newcommand{\indicator}[4]{\mathbbm{1}(#1_{#2}=#3_{#4})}
\newcommand{\mathbfcal}[1]{\bm{\mathcal{#1}}}
\newcommand{\real}{\mathbbm{R}}
\newcommand{\integer}{\mathbbm{Z}}
\newcolumntype{P}[1]{>{\centering\arraybackslash}p{#1}}

\newcommand{\appendixref}[1]{%
  \hyperref[#1]{Appendix~\ref*{#1}}\unskip
}

\title{Unsupervised deep learning for semantic segmentation of multispectral LiDAR forest point clouds}

\usepackage{authblk}

\author[1,\small\ensuremath{\ast}]{Lassi Ruoppa}
\author[1]{Oona Oinonen}
\author[1]{Josef Taher}
\author[1]{Matti Lehtomäki}
\author[2]{Narges Takhtkeshha}
\author[1]{Antero Kukko}
\author[1]{Harri Kaartinen}
\author[1,3]{Juha Hyyppä}

\affil[1]{Department of Remote Sensing and Photogrammetry, Finnish Geospatial Research Institute FGI, The National Land Survey of Finland, Vuorimiehentie 5, Espoo, FI-02150, Finland}
\affil[2]{3D Optical Metrology (3DOM) unit, Bruno Kessler Foundation (FBK), Trento, Italy}
\affil[3]{Department of Built Environment, Aalto University, School of Engineering, P.O. Box 11000, Aalto, FI-00076, Finland}

\begin{document}

\twocolumn[\begin{@twocolumnfalse}

\maketitle

\begin{abstract}
Point clouds captured with laser scanning systems from forest environments can be utilized in a wide variety of applications within forestry and plant ecology, such as the estimation of tree stem attributes, leaf angle distribution, and above-ground biomass. However, effectively utilizing the data in such tasks requires the semantic segmentation of the data into wood and foliage points, also known as leaf--wood separation. The traditional approach to leaf--wood separation has been geometry- and radiometry-based unsupervised algorithms, which tend to perform poorly on data captured with airborne laser scanning (ALS) systems, even with a high point density ($>1,000$ points/m$^2$). While recent machine and deep learning approaches achieve great results even on sparse point clouds, they require manually labeled training data, which is often extremely laborious to produce. Multispectral (MS) information has been demonstrated to have potential for improving the accuracy of leaf--wood separation, but quantitative assessment of its effects has been lacking. This study proposes a fully unsupervised deep learning method, GrowSP-ForMS, which is specifically designed for leaf--wood separation of high-density MS ALS point clouds (acquired with wavelengths 532, 905 and 1550 nm) and based on the GrowSP architecture. GrowSP-ForMS achieved a mean accuracy of 84.3\% and a mean intersection over union (mIoU) of 69.6\% on our MS test set, outperforming the unsupervised reference methods by a significant margin. When compared to supervised deep learning methods, our model performed similarly to the slightly older PointNet architecture but was outclassed by more recent approaches. Finally, two ablation studies were conducted, which demonstrated that our proposed changes increased the test set mIoU of GrowSP-ForMS by 29.4 percentage points (pp) in comparison to the original GrowSP model and that utilizing MS data improved the mIoU by 5.6 pp from the monospectral case.\\
\end{abstract}


\vspace{0.5cm}

\end{@twocolumnfalse}]

\renewcommand{\thefootnote}{\small\ensuremath{\ast}}
\footnotetext[1]{Corresponding author: Lassi Ruoppa (lassi.ruoppa@nls.fi)}
\renewcommand{\thefootnote}{\arabic{footnote}}
\setcounter{footnote}{0}

\section{Introduction} \label{section:introduction}

Three-dimensional (3D) point clouds acquired with light detection and ranging (LiDAR) scanners have a wide variety of applications within forest inventory and plant ecology, including estimation of tree stem attributes, leaf angle distribution, and above-ground biomass (AGB) \citep{wang2020unsupervised}. A majority of these applications require the semantic segmentation of each point into either wood or foliage, which is often referred to as \emph{leaf--wood separation}.

Traditionally, leaf--wood separation has been performed using \emph{unsupervised algorithms} that utilize either geometric or radiometric properties of the data and do not require any ground truth labels. Such approaches are generally designed for point clouds captured with stationary terrestrial laser scanning (TLS) systems that store tree structures with millimeter-level details. As such, the performance of the algorithms generally begins to decline as the density of the point cloud decreases \citep{vicari2019leaf,tian2022graph}. Consequently, while utilizing airborne laser scanning (ALS) would allow for the survey of larger areas spanning multiple square kilometers much more efficiently at the cost of reduced point density, conventional leaf--wood separation algorithms often struggle with even high density ALS data ($>1,000$ points/m$^2$) due to their heavy reliance on detailed point cloud geometry.

Following the recent successes of deep learning (DL) in fields such as computer vision and natural language processing, the amount of research on DL-based leaf--wood separation approaches has exploded \citep[see e.g.][]{xi2018filtering,krisanski2021sensor,kaijaluoto2022semantic,wielgosz2023point2tree,xiang2024automated}. Deep learning approaches seemingly do not suffer from the same problems as conventional leaf--wood separation algorithms when it comes to ALS data, as evidenced by the success of e.g. \cite{windrim2019forest}, \cite{windrim2020detection} and \cite{xiang2024automated}. Although DL-based methods have achieved extremely high accuracies in the task of leaf--wood separation, they require large amounts of manually generated training data with pointwise semantic annotations, which are both time-consuming and challenging to produce. In fact, according to a recent review, acquiring sufficient amounts of representative and labeled training data remains one of the main challenges of deep learning applications for forest inventory and planning \citep{hamedianfar2022deep}.

The prospect of an unsupervised deep learning method for leaf--wood separation is extremely appealing, as it would not require manually annotating data, while simultaneously being less reliant on detailed point cloud geometry. A number of unsupervised deep learning approaches for point cloud semantic segmentation have been proposed in recent years \citep{zhang2023growsp,chen2023pointdc,liu2024u3ds} and while these works have demonstrated great potential with performance approaching that of some early fully supervised semantic segmentation models, the difference in accuracy in comparison to state-of-the-art DL approaches remains fairly significant.

Conventional LiDAR scanners only capture intensity information from one wavelength. In recent years, multispectral (MS) laser scanning systems that operate on multiple distinct wavelengths have emerged as a promising and rapidly developing alternative \citep{takhtkesha2024multispectral}. Multiple works have demonstrated that utilizing multispectral point clouds improves results over conventional monospectral data in various forestry and ecology-related tasks, such as individual tree segmentation \citep[see e.g.][]{dai2018new,huo2020individual}, tree species classification \citep[see e.g.][]{yu2017single,kukkonen2019multispectral,hakula2023individual}, and structure and physiology measurements \citep{woodhouse2011multispectral}. Furthermore, \cite{li2013separating} demonstrated that multispectral data also has the potential for improving the accuracy of radiometry-based leaf--wood separation. Consequently, the auxiliary intensity information provided by MS systems could prove useful in unsupervised DL-based leaf--wood separation. However, to the best of our knowledge, the effect of multispectral information on leaf-wood segmentation accuracy from point clouds has not been quantitatively evaluated and compared to the monospectral case in any study.

In this work, we apply unsupervised deep learning for semantic segmentation of multispectral ALS forest data for leaf--wood separation on both coniferous and deciduous trees in the boreal forest zone. Our contributions are as follows:
\begin{enumerate}
    \item We introduce the first fully unsupervised deep learning approach for semantic segmentation of multispectral ALS forest point clouds. To our knowledge, unsupervised deep learning has previously not been applied to leaf--wood separation of any kind.
    \item Our proposed model, \textbf{GrowSP-ForMS}, is adapted from the pioneering GrowSP \citep{zhang2023growsp} architecture by introducing changes specifically designed for leaf--wood separation. By comparing the segmentation results of GrowSP-ForMS with unsupervised and supervised reference methods, we show that GrowSP-ForMS sets the new state-of-the-art in the unsupervised leaf--wood separation of MS ALS data and is comparable to an early fully supervised approach. Furthermore, by conducting an ablation study, we demonstrate that the proposed changes to GrowSP improve leaf--wood separation accuracy.
    \item We complement the pioneering studies of \cite{li2013separating} and \cite{li2018utilization} by reporting the first quantitative accuracy evaluations of leaf-wood separation comparing multispectral and monospectral point cloud data. An ablation study where GrowSP-ForMS is trained for each reflectance channel separately, as well as all two-channel combinations shows that utilizing multispectral data improves the segmentation accuracy in comparison to monospectral data.
\end{enumerate}

\section{Related work} \label{section:related_work}

\subsection{Deep-learning-based point cloud semantic segmentation}

\subsubsection{Fully supervised methods}

Some of the first deep learning architectures for semantic segmentation of point clouds discretized the inputs by projecting the point cloud onto a plane from several perspectives and subsequently utilized 2D convolutional neural networks (CNNs) \citep{su2015multiview,boulch2018snapnet}. Later, multiple works, such as SqueezeSeg and its successors, proposed using spherical projections to alleviate the problem of occlusions and defects present in multi-view projections \citep{wu2018squeezeseg,wu2019squeezesegv2,milioto2019rangenet++,xu2020squeezesegv3}.

Architectures such as SegCloud \citep{tchapmi2017segcloud} and PointGrid \citep{le2018pointgrid} first voxelize the input clouds and then perform semantic segmentation with a 3D CNN. Since the cubical scaling of the memory footprint of 3D convolutions poses a significant limitation in terms of both voxel size and depth of the networks, \cite{graham20183d} proposed using submanifold sparse convolutions (SSCs) for semantic segmentation of point clouds. SSCs significantly decreased the computational load in comparison to 3D CNNs, while simultaneously achieving state-of-the-art performance. \cite{choy20194d} further generalized SSCs for generic input and output coordinates as well as arbitrary kernel shapes.

Besides voxelization, a variety of more complex discretization approaches that exploit the inherent sparsity of point clouds have been proposed. Examples include OctNet \citep{riegler2017octnet} and O-CNN \citep{wang2017ocnn}, which both utilize novel octree structures, as well as SPLATNet \citep{su2018splatnet} and LatticeNet \citep{rosu2019latticenet}, which opt for lattice-based data structures.

PointNet \citep{qi2017pointnet} was the first neural network capable of processing raw point clouds directly. However, PointNets inability to capture local geometric information hindered its performance in semantic segmentation tasks. To this end, the authors introduced an improved version, PointNet++, which utilized a hierarchical structure for extracting local features from small point neighborhoods  \citep{qi2017pointnet++}. \cite{hu2020randlanet} proposed RandLA-Net for efficient semantic segmentation of large-scale point clouds. The network achieved significant performance gains through an innovative use of random downsampling and a novel local feature aggregation module.

Several works have proposed pointwise convolutions as an alternative to applying CNNs to discretized point clouds. PointCNN \citep{li2018pointcnn} introduced the $\bm{\chi}$-convolution which leveraged spatially-local correlation from unstructured point cloud data, while KPConv \citep{thomas2019kpconv} used a set of kernel points to determine where each convolution kernel weight is applied. On the other hand, since point clouds can easily be transformed into graph representations, graph convolutional neural networks (GCNN) have also been a popular approach for semantic segmentation. SPG \citep{landrieu2018large} combined GCNNs with superpoint representations, while DGCNN \citep{wang2019dynamic} and SPH3D \citep{lei2021spherical} both introduced novel graph convolutional operators designed specifically for point clouds.

More recently, following the success of transformer-based architectures in natural language processing and computer vision tasks, a number of works have proposed architectures adapted for semantic segmentation of point cloud data \citep{guo2021pct,zhao2021point,wu2022point,wu2024point,robert2023efficient}.

\subsubsection{Weakly supervised methods}

Manually generating ground truth labels for large amounts of data is extremely time-consuming, especially when it comes to data formats that are challenging to annotate, like 3D point clouds. Weakly supervised deep learning attempts to alleviate this problem by training models with a very limited amount of ground truth data.

Since manually generating semantic annotations for 2D images requires significantly less labeling than 3D point clouds, utilizing 2D ground truth labels for 3D deep learning is a form of weak supervision, which some works have explored \citep{wang2019towards,kweon2022joint}. However, the more common strategy for weak supervision is using a limited amount of ground truth labels, such that only a small minority of all available data with annotations is used for training. In principle, training such a model would only require manually annotating a small subset of points. \cite{xu2020weakly} were among the first to apply this form of weak supervision in the context of point cloud semantic segmentation. By combining a 4-part loss function with label propagation at the inference stage, their method achieved an accuracy comparable to that of supervised baselines with only 10\% of the labels. PSD \citep{zhang2021perturbed} with its perturbed self-distillation framework achieved performance comparable to RandLA-Net using 1\% of available point annotations, while SQN \citep{hu2022sqn} introduced a point feature query network which enabled back-propagating training signals from sparsely annotated points to a wider context, resulting in performance comparable to supervised baselines on multiple benchmark data sets using just 0.1\% of ground truth labels. Transformer architectures have also been adapted to weakly supervised point cloud semantic segmentation by \cite{yang2022mil}. Contrastive scene contexts proposed a slightly different approach for training semantic segmentation networks in a weakly supervised manner by introducing an unsupervised contrastive pretraining phase \citep{hou2021exploring}.

In contrast to methods that only utilize a limited amount of labeled data for training, pseudo-label-based methods assign artificial labels for unlabeled data points. SSPC-Net \citep{chen2021sspcnet} and HybridCR \citep{li2022hybridcr} both presented weakly supervised point cloud semantic segmentation frameworks based on pseudo-labels. The former utilized a combination of superpoints and dynamic label propagation, while the latter imposed a contrastive loss on pseudo-labels generated from backbone predictions. LESS \citep{liu2022less} presented a novel strategy for generating sparsely annotated data, where the labels are divided into three separate categories during training sparse, weak, and propagated, each of which has a separate contrastive loss function.

\subsubsection{Unsupervised methods}

Research on unsupervised deep learning approaches for semantic segmentation of point clouds remains fairly limited. The few existing architectures all utilize some form of deep clustering, where extracted feature vectors are clustered and subsequently used as pseudo-labels, which are iteratively updated during training.

GrowSP \citep{zhang2023growsp} first oversegments the input point clouds into superpoints and subsequently constructs superpoint level feature representations based on pointwise feature vectors from a sparse convolutional neural network (SCNN) feature extractor. The features are then clustered with $k$-means to obtain pseudo-labels for training the network. U3DS$^{3}$ \citep{liu2024u3ds} and PointDC \citep{chen2023pointdc} both utilize a very similar framework based on oversegmented point clouds and a 3D CNN backbone. To improve feature robustness, U3DS$^{3}$ used two separate pathways in the feature extractor combined with a two-part loss function. On the other hand, PointDC utilizes cross-modal distillation (CMD), where the input point clouds are first converted into multi-view images and passed as input to a pretrained unsupervised 2D feature extractor. The 2D features are subsequently projected back into 3D and used as supervision for training the feature extractor network.

Perhaps unsurprisingly, the performance of 3D unsupervised approaches is far from state-of-the-art supervised and weakly supervised methods in terms of segmentation accuracy. However, models such as GrowSP and U3DS$^{3}$ both show great potential and achieve results that are comparable to early supervised architectures such as PointNet \citep{zhang2023growsp,liu2024u3ds}.

\subsection{Multispectral laser scanning}

Multispectral LiDAR represents the next generation of laser scanning and has been shown to be beneficial in various applications related to vegetation and object classification \citep{kaasalainen2007toward, kaasalainen2019multispectral}. MS representations provide a large number of spectral features, which have the potential to yield improvements in automatic classification and segmentation of point clouds \citep{kaasalainen2007toward}. Forestry and ecology have been some of the most popular applications for utilizing multispectral data \citep{takhtkesha2024multispectral} and its feasibility has been demonstrated in a wide variety of tasks, including individual tree segmentation \citep[see e.g.][]{dai2018new,huo2020individual}, tree species classification \cite[see e.g][]{yu2017single,budei2018identifying,lindeberg2021classification}, stem volume estimation \citep{axelsson2023use}, leaf--wood separation \citep{li2013separating,howe2015capabilities,li2018utilization}, forest environment classification \citep{hopkinson2016multisensor}, and structure and physiology measurements in forest ecosystems \citep{woodhouse2011multispectral}. Notably, \cite{yu2017single}, \cite{kukkonen2019multispectral}, and \cite{hakula2023individual} all found that multispectral data improved the classification accuracy of individual tree species in boreal forests in comparison to monospectral data. Similarly, \cite{dai2018new} and \cite{huo2020individual} both established that multispectral data significantly improved detection accuracy in individual tree segmentation.

Outside of forestry applications, multispectral data has been used for a wide variety of remote sensing tasks, including change detection \citep{matikainen2017object,matikainen2019toward}, road mapping \citep{karila2017feasibility}, and land cover classification \citep{wang2014airborne,wichmann2015evaluating,bakula2016testing,teo2017analysis}, which appears to be the most popular application. Similarly to the case of tree species classification, several works have demonstrated that utilizing intensity information from multiple wavelengths improves the accuracy of land cover classifiers \citep{wang2014airborne,matikainen2017object,teo2017analysis}. More recent research efforts on land cover classification using multispectral data have focused on deep learning approaches \citep{yu2020hybrid,pan2020landcover,wang2021multi,li2022agfpnet,zhang2022introducing}, including weakly supervised methods \citep{chen2024feature,takhtkeshha2024automatic}. \cite{oinonen2024unsupervised} proposed GroupSP, a fully unsupervised DL framework based on GrowSP designed for semantic segmentation of multispectral point clouds depicting urban environments.

\subsection{Leaf--wood separation}

\subsubsection{Unsupervised algorithms}

Traditional leaf--wood separation algorithms can broadly be divided into two classes, radiometry-based methods and methods based on point cloud geometry \citep{wang2020lewos}. Such algorithms are unsupervised, since neither approach requires any prior knowledge about the classes of individual points, rather, the classification is based on heuristic assumptions about the geometric or radiometric properties of wood points, as well as the general structure of trees.

\cite{cote2009structural} and \cite{beland2014seeing} both extracted wood points from TLS scans based on two fixed thresholds determined through careful data analysis. On the other hand, \cite{wu20133d} extracted wood points from simulated full waveform LiDAR using a single intensity threshold, which was optimized by applying the density-based spatial clustering of applications with noise (DBSCAN) \citep{ester1996density} algorithm to leaf-off point clouds. Similarly, \cite{yang2013three} detected wood points from full waveform TLS scans with a single threshold based on the distribution of reflectance value and pulse width ratios. By merging point clouds from near-infrared and shortwave-infrared scanners, \cite{li2013separating} demonstrated the potential for improving radiometry-based classification of wood and foliage points through the use of multispectral data. Radiometry-based methods have fallen out of favor in more recent research due to a few notable limitations, such as external factors influencing the optical properties that thresholding relies on \citep{wang2020lewos} and differences in attributes of different scanners and tree species that limit the generalization of thresholds to other types of data \citep{tao2015geometric}.

Early geometry-based leaf--wood separation algorithms were generally designed for the purpose of 3D reconstruction of trees, rather than semantic segmentation. \cite{xu2007knowledge} separated wood points starting from predefined root points using Dijkstra's shortest path algorithm, while \cite{livny2010automatic} extracted initial branch structure graphs with multi-root Dijkstra’s algorithm. On the other hand, \cite{raumonen2013fast} proposed a local approach for constructing precision tree models from TLS scans where the wooden components are iteratively reconstructed from small connected surface patches. \cite{belton2013processing} clustered geometric descriptors of point clouds representing individual trees with a Gaussian mixture model (GMM) and subsequently manually classified the clusters into trunk, branches, and foliage.

\cite{tao2015geometric} presented one of the first geometry-based algorithms designed for leaf--wood separation from point clouds of individual trees. Their approach detects initial wood proposals using thresholds for fitted circles and lines, which is followed by Dijkstra's shortest path algorithm and range search. \cite{wang2018separating} proposed a leaf--wood separation method for entire forest scenes, where wood points are detected through dynamic segment merging of superpoints and linearity thresholding.

A significant number of geometric leaf--wood separation algorithms are based on modeling the point cloud as a graph. LeWoS \citep{wang2020lewos} estimates the wood probability of connected components in the point cloud graph and subsequently applies regularization to form the final prediction. \cite{wang2020unsupervised} later improved the method with additional geometric features and graph optimization after the initial wood probability estimation. On the other hand, GBS \citep{tian2022graph} performs leaf--wood separation on graph representations of point clouds depicting individual trees using a combination of shortest path analysis, multi-scale thresholding based on geometric descriptors, and region growing. The algorithm of \cite{vicari2019leaf} combines GMM clustering and graph-segmentation by utilizing four separate classification branches, which are then merged into a single class prediction for each point.

Outside of graph-based approaches, \cite{wan2021novel} classified wood and foliage using local point cloud curvature and connected components. \cite{shcherbcheva2023unsupervised} proposed a statistical approach, which first fits a GMM to a set of geometric point cloud descriptors and subsequently detects inflection points between the resulting centroids and utilizes them as flexible thresholds for discerning between leaf and wood points.

\subsubsection{Machine- and deep-learning-based approaches}

A wide variety of fully supervised machine- and deep-learning-based approaches have been proposed for leaf--wood separation. Supervised methods have two advantages over unsupervised algorithms: they tend to be more accurate \citep[see e.g.][]{morel2020segmentation,jiang2023lwsnet} and generally require considerably less hyperparameter optimization. However, the obvious drawback of utilizing supervised learning is the requirement for manually generated pointwise ground truth labels. To the best of our knowledge, neither weakly supervised nor unsupervised learning has been applied to leaf--wood separation in any previous work.

\cite{ma2016improved} introduced GAFPC for supervised leaf--wood separation, which combines GMMs with six consecutive filters designed to correct misclassified points and remove noise. \cite{yun2016novel} trained a support vector machine (SVM) to separate wood and foliage points in TLS scans of individual deciduous trees, achieving an overall accuracy of over 90\% for most species. To assess the feasibility of supervised machine learning for leaf--wood separation, \cite{wang2017feasibility} conducted a performance comparison between SVM, naive Bayes classifier, GMM, and random forest. Although a majority of the approaches reached an overall accuracy of at least 95\%, random forest proved the most accurate.

\cite{li2018utilization} used a random forest classifier for separating leaf and wood points from dual-wavelength full waveform TLS point clouds, while \cite{zhu2018foliar} trained a random forest classifier for leaf--wood separation of RGB colorized TLS data with optimal geometric and radiometric features obtained using adaptive radius near-neighbor search. Similarly, \cite{zhou2019separating} combined random forest classifiers with a novel multi-scale strategy, where geometric features are computed for several neighborhood sizes. \cite{moorthy2020improved} compared the leaf--wood separation accuracy of random forest and two gradient boosting methods using similar multi-scale features, with random forest yielding the best performance.

Point cloud CNNs have been one of the most popular approaches for DL-based semantic segmentation of forest data. One of the earliest works to apply deep learning to the task of leaf--wood separation utilized a 3D fully convolutional neural network for semantic segmentation of TLS forest point clouds, processing the inputs as $128\times128\times128$ blocks of uniform voxels \citep{xi2018filtering}. \citet{windrim2019forest} proposed a pipeline for segmenting ALS forest data, where individual trees were first delineated from rasterized point clouds using Faster R-CNN \citep{ren2017faster} followed by semantic segmentation with a 3D CNN inspired by VoxNet \citep{maturana2015voxnet}. The performance of the pipeline was further improved in later work with the inclusion of LiDAR intensity as one of the input features for the semantic segmentation network \citep{windrim2020detection}. \cite{shen2022deep} introduced a slightly different approach, where TLS point clouds are first partitioned into geometrically consistent regions, after which under-represented geometric features are compensated with a novel feature balance module. Finally, the balanced data is used for training a PointCNN semantic segmentation network.

In addition to CNNs, both the standard PointNet++ architecture and modified variants have been applied for leaf--wood separation in multiple works. \cite{kim2023automated} and \cite{wielgosz2023point2tree} both utilized a standard PointNet++ model for semantic segmentation of forest TLS point clouds, while \citet{morel2020segmentation} combined geometric leaf--wood separation algorithms with a modified PointNet++ and trained it with simulated TLS point clouds of individual trees. On the other hand, \cite{krisanski2021sensor} proposed a sensor-agnostic semantic segmentation model based on a modified PointNet++ backbone trained on a diverse data set consisting of point clouds captured with various types of laser scanners.

\cite{kaijaluoto2022semantic} presented a novel approach for semantic segmentation of MLS forest point clouds using raw (non-georeferenced) 2D laser scanner profiles as a series of 2D rasters, each of which contained the ranges, reflectances and echo deviations measured during a single scanner mirror rotation. Their U-Net-based 2D CNN achieved performance comparable with RandLA-Net at a speed sufficient for real-time applications.

While a majority of research efforts have been focused on data preprocessing and feature engineering, more novel semantic segmentation architectures for leaf--wood separation have also been proposed. LWSNet \citep{jiang2023lwsnet} augmented a standard 3D U-Net with local contextual feature enhancement via a rearrangement attention mechanism and residual connection optimization, while MDC-Net \citep{dai2023mdcnet} introduced a novel multi-directional collaborative convolutional module designed for extracting discriminative features for leaf--wood separation and combined it with a multi-scale 3D CNN.

More recently, \cite{xiang2024automated} proposed ForAINet for panoptic segmentation of forest point clouds. The architecture utilizes a 3D SCNN U-Net for feature extraction followed by three separate branches for semantic prediction, center offset prediction, and feature embedding.

\section{Material and methods} \label{section:material_and_methods}

\subsection{Data set}

While there are some openly available point cloud data sets for semantic segmentation of forests \citep{momo2018data,wang2020lewosdata,kaijaluoto2022semantic,puliti2023forinstance}, none of them are multispectral. In order to benchmark our proposed leaf--wood separation approach, we composed a novel high-density multispectral point cloud data set captured in a boreal forest environment. In addition, we manually generated semantic labels for an 8 million point subsection of the data set.

\subsubsection{Data acquisition} \label{section:data_acquisition}

We utilized the same raw data set as \cite{hakula2023individual}, which was captured 22nd of June 2021 using the Finnish Geospatial Research Institute's (FGI's) in-house developed laser scanner system HeliALS-TW. The system comprises three different Riegl LiDAR scanners (Riegl GmbH, Austria) VUX-1HA, miniVUX-3UAV, and VQ-840-G, which we refer to as scanners 1, 2, and 3 respectively. A summary of notable scanner-specific attributes is provided in \autoref{table:scanners}. To track its position and orientation, HeliALS-TW is equipped with a GNSS-IMU positioning system consisting of a NovAtel ISA-100C inertial measurement unit (IMU), a NovAtel PwrPak7 global navigation satellite system (GNSS) receiver, and a NovAtel GNSS-850 antenna \citep{hakula2023individual}. Scanners 1 and 2 providing cross-track profiles were installed to scan at 15 degrees forward tilt angle, while scanner 3 operates using a conical scanning at 40 degree cross-track and 28 degrees angle in flight direction.

\begin{table*}[h!]
    \centering
    \caption{Technical specifications of the three separate LiDAR scanners in the multispectral HeliALS-TW system. The beam divergence and diameter of scanner 2 are expressed as two values due to its elliptical beam and the maximum scanning angle of scanner 3 is expressed as \emph{angle in flight direction} $\times$ \emph{the angle perpendicular to flight direction}. The reported approximate point density is greater than the number of sent pulses since a single pulse will frequently produce multiple returns in forested areas. The table is reproduced based on \cite{hakula2023individual}, where the same system was used.}
    \bigskip
    \begin{tabular}{llll}
        \toprule
        \textbf{Scanner} & \textbf{1} & \textbf{2} & \textbf{3} \\ \midrule \midrule
        \textbf{Model} & VUX-1HA & miniVUX-3UAV & VQ-840-G \\
        \textbf{Wavelength (nm)} & 1,550 & 905 & 532 \\
        \textbf{Approximate flight altitude (m)} & 80 & 80 & 80 \\
        \textbf{Approximate point density (points/m$^2$)} & 1,400 & 500 & 1,600 \\
        \textbf{Maximum number of returns} & 12 & 5 & 15 \\
        \textbf{Maximum scanning angle ($^\circ$)} & 360 & 120 & $28\times40$ \\
        \textbf{Laser beam divergence (mrad)} & 0.5 & $0.5\times1.6$ & 1 \\
        \textbf{Approximate footprint (nadir) at ground level (cm)} & 4 & $4\times12.8$ & 8 \\
        \textbf{Pulse repetition rate (kHz)} & 1,017 & 300 & 200 \\ \bottomrule
    \end{tabular}
    \label{table:scanners}
\end{table*}

The data set was collected in the Scan Forest test site near Evo, Finland ($61.19^\circ$N, $25.11^\circ$E). To capture the data, a helicopter equipped with the HeliALS-TW system was flown over multiple predefined circular test sites with 55 m diameter from two perpendicular directions at an approximate altitude of 80 m above ground level. This study utilizes point clouds characterizing boreal forest environments from a total of 20 test sites.

The trajectory of HeliALS-TW was computed with Waypoint Inertial Explorer \citep[version 8.90,][]{novatel2022waypoint} and a single virtual GNSS base station from Trimnet service (RINEX 3.04) positioned approximately at the centroid of the Evo research forest area (maximum baseline $<4$ km). The raw individual scans were subsequently georeferenced in the Riegl RiPROCESS software by utilizing the GNSS and IMU data. Finally, the georeferenced respective monospectral point clouds from scanners 1, 2, and 3 were combined into a multispectral point cloud using $k$-nearest neighbors interpolation with $k=1$. For points that had no neighbors from one or more scanners within a threshold distance of 0.25 m, the neighbors were considered missing and the corresponding reflectance fields were left empty. A visualization of one MS point cloud depicting a circular test site is shown in \autoref{fig:training_data_vis} (a).

\subsubsection{Point cloud normalization and ground removal}

\begin{figure*}[!ht]
\centering
\includegraphics[width=0.9\textwidth]{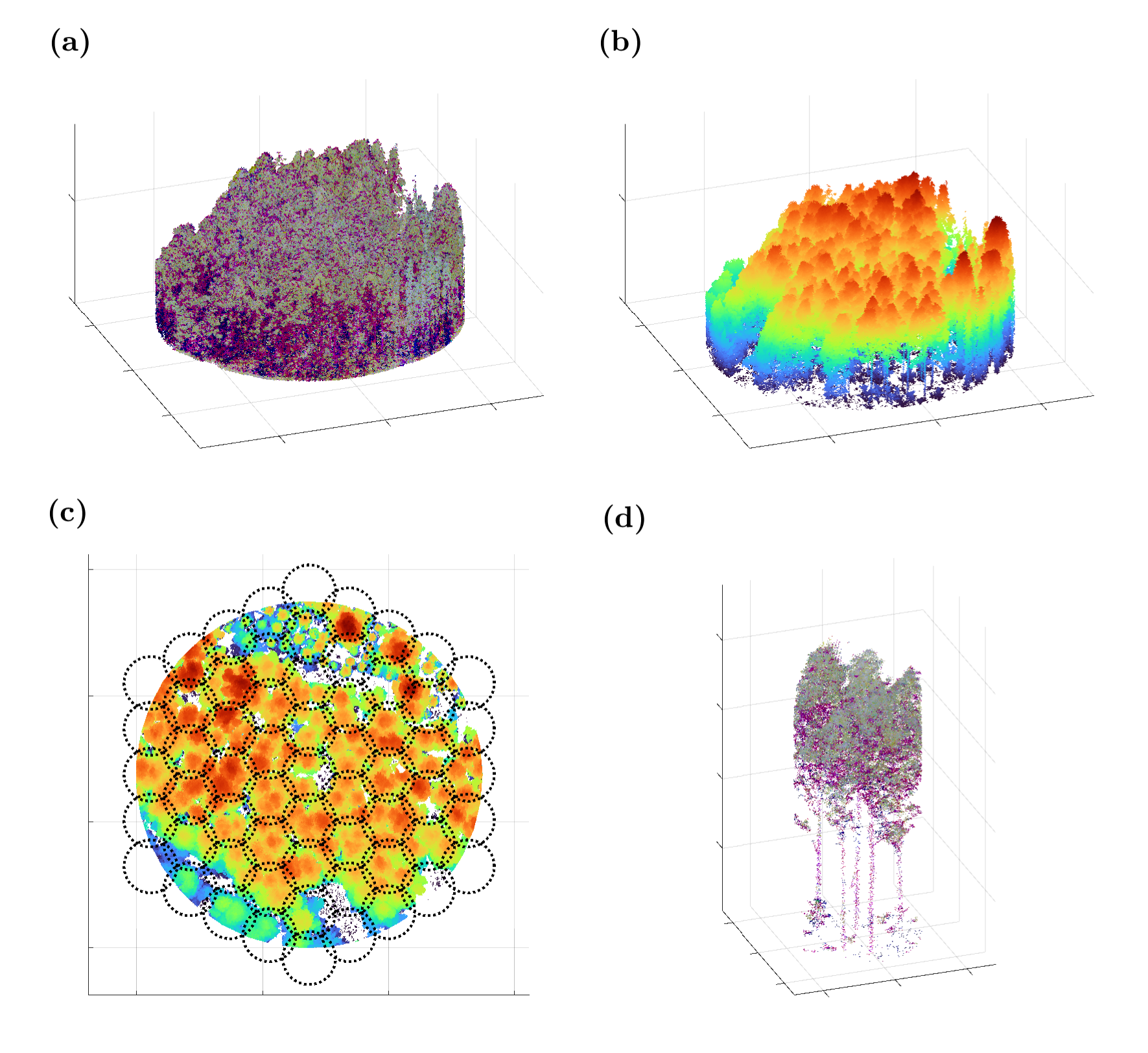}
\caption{Visualization of the data preprocessing steps for plot \#2. The points in (b) and (c) are colored based on the $z$-coordinates, while (a) and (d) use pseudo colors generated from scaled reflectance values of scanners 1, 2, and 3 for the red, green, and blue channels respectively. (a) Original multispectral point cloud of the forest plot. (b) Normalized point cloud from which the ground points and elevation have been removed. (c) Outlines of the cylindrical neighborhoods of radius $r_c=4.2$ m used as training data superimposed on the normalized point cloud in the $xy$-plane. (d) Example of a cylindrical neighborhood that has been cut out from the normalized point cloud.}
\label{fig:training_data_vis}
\end{figure*}

The point clouds of each test site were normalized by subtracting the digital terrain model (DTM) based on estimated ground elevation from the $z$-coordinates. Following \cite{hakula2023individual}, the DTM was reconstructed from smooth surface patches extracted from the point cloud by combining the lowest point search of \cite{hyyppa2020accurate} and the ground extraction algorithm of \cite{lehtomäki2016object}. Subsequent to normalizing, the detected ground points were removed from the test site point clouds. A comparison between an original test site point cloud and the normalized version with the ground removed is shown in \autoref{fig:training_data_vis} (a)--(b).

\subsubsection{Manual annotation of data}

Although training an unsupervised model does not require manually annotated ground truth labels, we generated a limited amount of labeled data comprised of two test sites, denoted by \#1 and \#2, in order to quantitatively evaluate model segmentation accuracy and objectively compare with fully supervised leaf--wood separation approaches on the same data. The two plots to label were chosen such that they comprehensively represent the overall data set, containing forests of different densities and all major tree species in the data, specifically Scots pine (\emph{Pinus sylvestris} L.), Norway spruce (\emph{Picea abies} (L.) H. Karst), Downy birch (\emph{Betula pubescens} Ehrh.) and Silver birch (\emph{Betula pendula} Roth). Plot \#1 is relatively sparse and consists of almost exclusively pine trees, while plot \#2 is significantly denser and consists primarily of birches and spruces.

While multiple different definitions for semantic ground truth classes exist within the context of forest data, we opted for the one most commonly used in the literature, where trees are split into wooden components and foliage. The division has been widely utilized in the context of machine learning \citep{zhou2019separating,wang2017feasibility}, deep learning \citep{shen2022deep,morel2020segmentation,krisanski2021sensor,jiang2023lwsnet,dai2023mdcnet} as well as geometry and radiometry based algorithms \citep{wang2018separating,wang2020lewos,wang2020unsupervised,tian2022graph}. Other less common definitions include further separating the wood components into stem and branch points \citep{puliti2023forinstance,kim2023automated} and only classifying the trunk as wood while considering branches foliage \citep{kaijaluoto2022semantic,windrim2020detection,wielgosz2023point2tree}.

Plots \#1 and \#2 were manually annotated using the point cloud processing tool CloudCompare \citep{girardeu2022cloudcompare} and a workflow similar to \cite{wielgosz2023point2tree}. The annotation process consisted of two steps: tree instance segmentation and semantic segmentation. In the first step, all sections that could reasonably be identified as a tree and at least 3 m tall were separated into individual segments. Since both plots contained partial trees around the perimeter, only trees with $\gtrsim50$\% of their points located within the test site boundaries were segmented. In the second step, each point of all tree instances from the previous step was classified as either wood or foliage. The former class consisted of stem and branch points, while the latter comprised all remaining points in the tree, such as leaves and needles. Finally, semantic segmentation was performed for the remaining points that were not considered part of a tree instance in the first step.

\begin{figure*}[!ht]
\centering
\includegraphics[width=\textwidth]{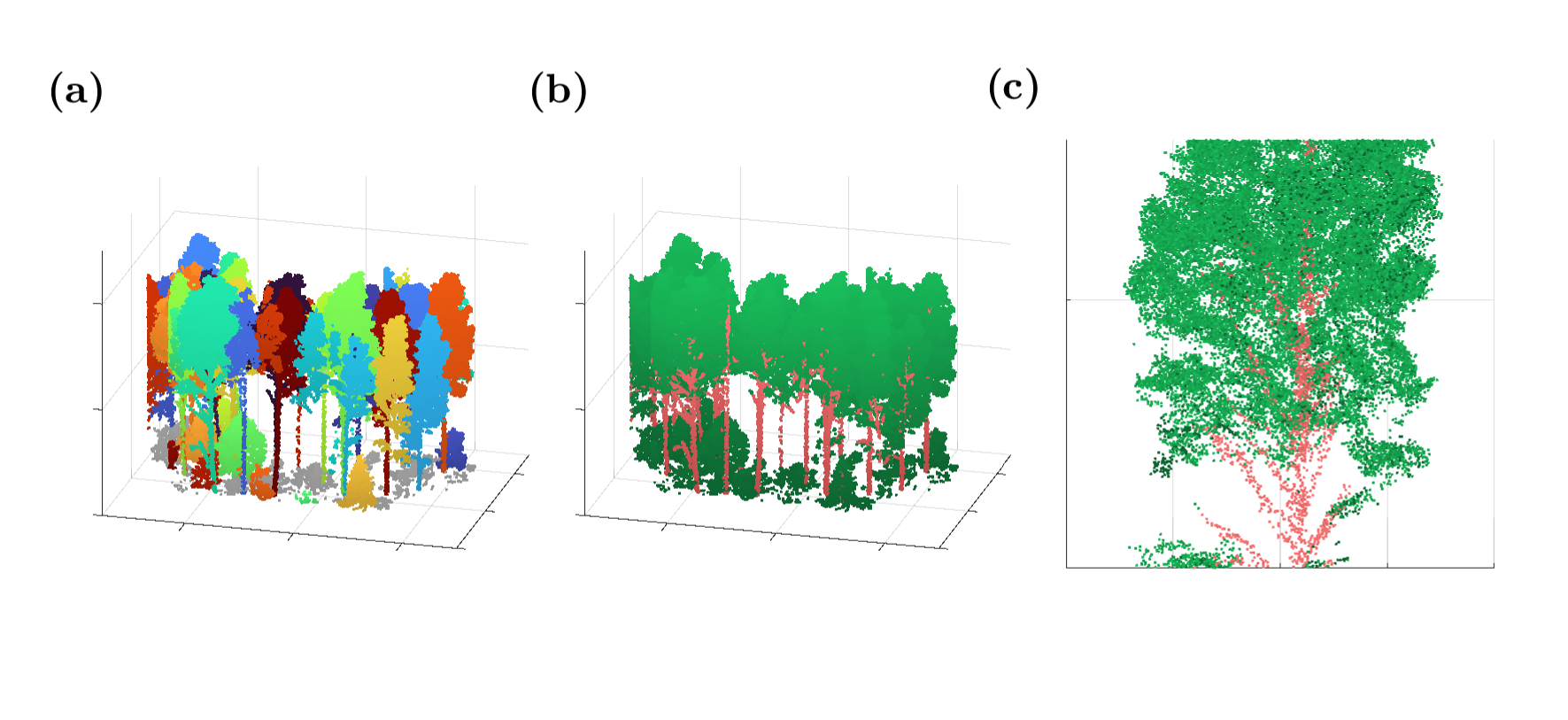}
\caption{Visualization of a section from the manually annotated point cloud of plot \#2. (a) Instance segmentation of trees, where each individual tree has been colored with a distinct color. Points that are not a part of any segmented tree instance have been colored gray. (b) Semantic annotations of individual points. The colors red and green represent wood and foliage points respectively. (c) Section of an individual tree showing the finer details of the semantic annotation.}
\label{fig:annotation_vis}
\end{figure*}

The entire annotation process took around 270 hours and was carried out by a single annotator. The labeled portion of our data set contains a total of over 8 million manually annotated points, around 400,000 of which are classified wood. The exact class distributions are shown in \autoref{table:annotation_distribution}, while an example of the annotated data is shown in \autoref{fig:annotation_vis}. Although the annotations are of high quality, some number of erroneously labeled points is to be expected, which is generally the case when manually generating semantic annotations for point clouds \citep{kaijaluoto2022semantic,puliti2023forinstance}.

\begin{table}[!ht]
    \centering
    \caption{Class distribution of the two manually annotated plots separately and as a combined data set. The notable imbalance between classes displayed in the distribution is a problem that is commonly experienced with real-world data.} \bigskip
    \begin{tabular}{rcc}
        \toprule
        \textbf{Data} & \textbf{\#Foliage points} & \textbf{\#Wood points} \\ \midrule \midrule
        Plot \#1 & 2,738,606 \textbf{(93.2\%)} & 198,670 \textbf{(6.8\%)} \\
        Plot \#2 & 5,112,640 \textbf{(96.0\%)} & 215,532 \textbf{(4.0\%)} \\
        All & 7,851,246 \textbf{(95.0\%)} & 414,202 \textbf{(5.0\%)} \\ \bottomrule
    \end{tabular}
    \label{table:annotation_distribution}
\end{table}

\subsubsection{Training and test data} \label{section:data_preprocessing}

\begin{figure*}[!ht]
\centering
\includegraphics[width=\textwidth]{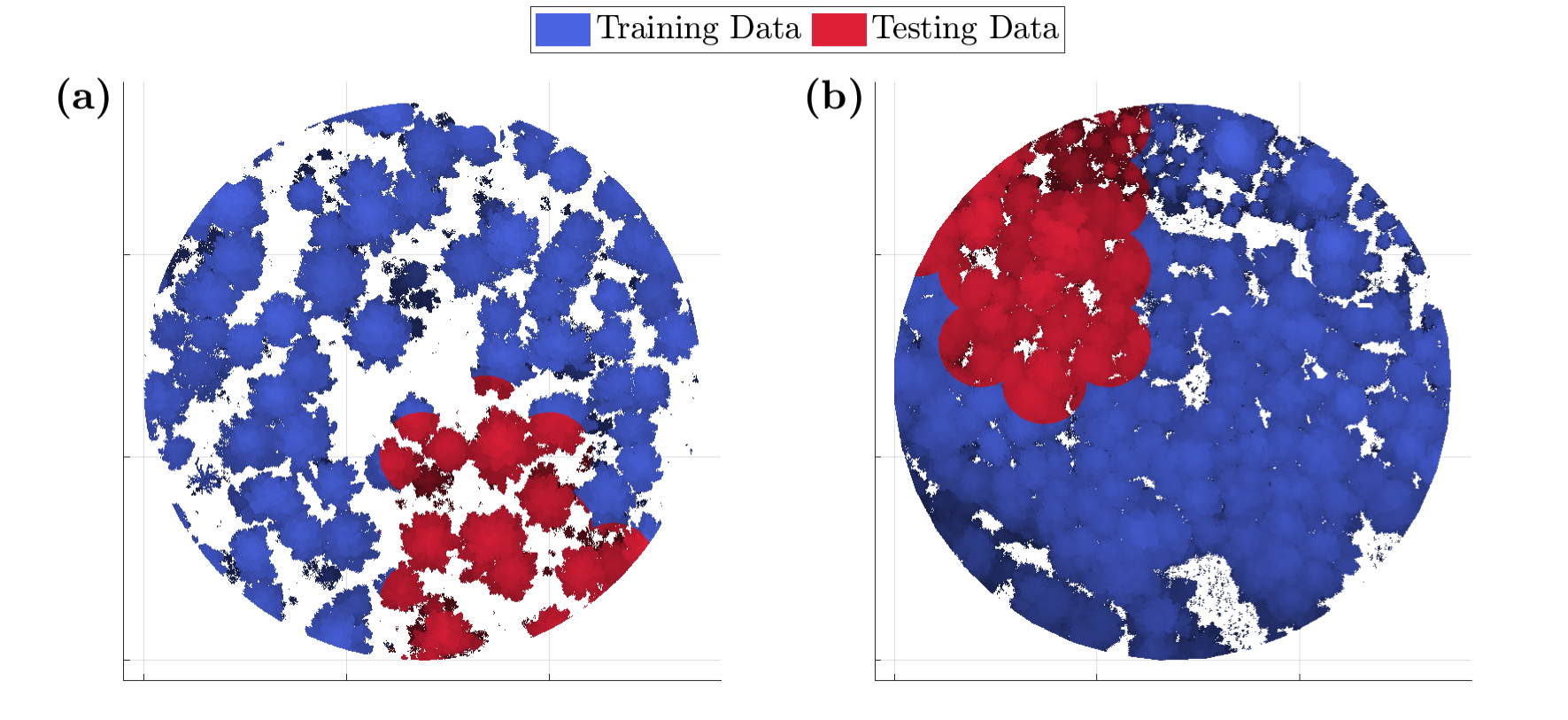}
\caption{Visualization of the split between training and testing data for our manually labeled data set. (a) Train-test split for plot \#1. (b) Train-test split for plot \#2.}
\label{fig:train_test_vis}
\end{figure*}

In order to train deep learning models, the full test site point clouds with a 55 m diameter were further divided into smaller cylindrical neighborhoods with a $r_c=4.2$ m radius, each cylinder containing 135,000 points on average. We opted to use cylindrical neighborhoods following \cite{xiang2023towards}, who found it the optimal neighborhood definition for panoptic segmentation of trees from the standpoint of computational efficiency and preserving vertical structures.

Since the circular test sites cannot be covered by smaller cylinders without a degree of overlap, we opted to arrange the cylinders in a hexagonal lattice (see \autoref{fig:training_data_vis} (c)) such that the area of overlap between adjacent cylinders was minimized. This was achieved by setting the $xy$-distance $d_c$ between the centers of adjacent cylinders to:
\begin{equation}
    d_c=\sqrt{3}r_c,
\end{equation}
which has been shown to be the most optimal method of covering a plane with circles in terms of minimizing the area of overlap \citep{kershner1939number}.

Not all points in the multispectral point cloud are guaranteed to contain reflectance values from all three scanners (see \autoref{section:data_acquisition}), which may pose a problem in the context of unsupervised learning: two points that are missing a reflectance value are not necessarily semantically similar, but the model may still group all such points into one class. In an attempt to alleviate the effects of missing reflectance values, we employ a simple strategy for populating the empty reflectance fields. Our unsupervised DL-based leaf--wood separation model initially segments the input point clouds into superpoints, which we employ as auxiliary information. Having constructed the superpoints using the algorithm detailed in \autoref{section:growsp_mods}, the missing reflectance values within each superpoint are populated with the mean of the available values on the corresponding channel. In the rare case where all reflectance values of a channel are empty within some superpoint, the mean reflectance value of the closest superpoint with reflectance data available is used instead.

The $xy$-coordinates within each cylinder were centered by subtracting the corresponding means from the original coordinate values. Similarly, the $z$-coordinates were normalized by subtracting the minimum $z$-value within the cylinder. Additionally, the reflectance features were normalized using an outlier robust normalization scheme. For each cylinder, given the reflectance features of channel $i$, denoted $\textbf{x}_{\text{reflectance}}^i$, we first computed the interquartile difference:
\begin{equation}
    \text{IQR}=Q_{75}(\textbf{x}_{\text{reflectance}}^i)-Q_{25}(\textbf{x}_{\text{reflectance}}^i),
\end{equation}
where $Q_j(\cdot)$ is the $j$th quantile function. The normalized reflectance feature vector $\Bar{\textbf{x}}_{\text{reflectance}}^i$ was subsequently computed as follows:
\begin{equation}
    \Bar{\textbf{x}}_{\text{reflectance}}^i=
    \frac{\textbf{x}_{\text{reflectance}}^i-M(\textbf{x}_{\text{reflectance}}^i)}{\text{IQR}}-
    \min(\textbf{x}_{\text{reflectance}}^i),
\end{equation}
where $M(\cdot)$ is the median of the input.

Since the fully supervised reference models required training data with ground truth labels, the annotated data set was divided into separate training and test splits. A continuous region of 12 cylinders was selected as the test set from each of the two annotated plots and the remaining $\sim80\%$ of the labeled cylinders were used for training. In its entirety, the test set is composed of around 2 million points. The exact division is visualized in \autoref{fig:train_test_vis}. Since the amount of manually labeled data was somewhat limited, the training data was not further separated into validation and training splits.

\subsection{Unsupervised deep learning model for leaf--wood separation}

We propose a novel unsupervised deep learning architecture for semantic segmentation of multispectral forest point clouds, referred to as \textbf{GrowSP-ForMS} (\textbf{GrowSP} for \textbf{for}est area \textbf{m}ulti\textbf{s}pectral point clouds). The model has been adapted from the GrowSP architecture \citep{zhang2023growsp} by introducing modifications specifically designed to improve segmentation accuracy on MS forest data. We first give a detailed overview of the original GrowSP architecture and subsequently present our proposed improvements.

\subsubsection{Original GrowSP architecture} \label{section:growsp_original}

\begin{figure*}[!ht]
\centering
\includegraphics[width=\textwidth]{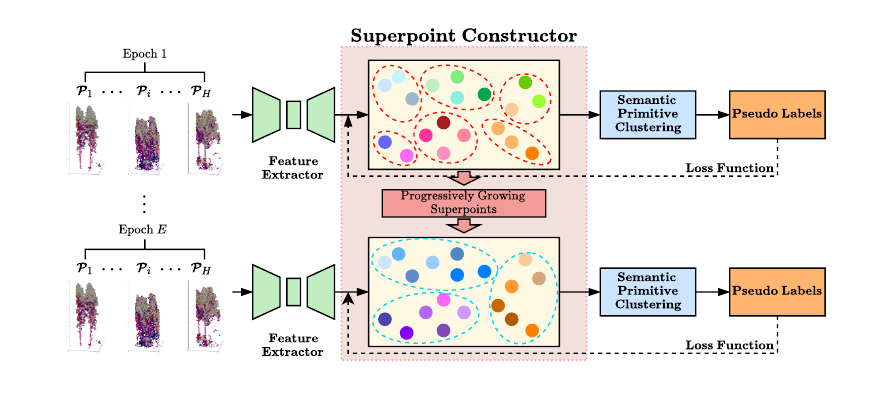}
\caption{Overview of the generic GrowSP model architecture. The learning framework consists of three main modules: a superpoint constructor, a neural network feature extractor, and a semantic primitive clustering module. In the figure, $\mathbfcal{P}_i$ denotes the $i$th 3D point cloud in a given data set. The figure has been reconstructed based on \cite{zhang2023growsp}.}
\label{fig:growsp_overview}
\end{figure*}

GrowSP approaches unsupervised semantic segmentation of point clouds as a joint problem of point cloud feature extraction and clustering. The architecture consists of three main modules: \emph{a superpoint constructor}, \emph{a neural network feature extractor}, and a \emph{semantic primitive clustering module}. A visual overview of the generic GrowSP model is shown in \autoref{fig:growsp_overview}.

The input of GrowSP is a data set $\mathbf{P}=\{\mathbfcal{P}_1,\ldots,\mathbfcal{P}_H\}$ consisting of $H$ point clouds each of which contains the $xyz$-coordinates of points and optional attributes such as color or LiDAR intensity. As a preprocessing step, GrowSP constructs initial superpoints for each input point cloud $\mathbfcal{P}_i\in\mathbf{P}$. Superpoints are segments of the original point clouds composed of multiple points that ideally belong to the same semantic category. Processing the point clouds on the superpoint level rather than point level reduces computational overhead and simplifies the process of extracting high-level semantic information since a superpoint is more likely to have intrinsic geometric meaning than individual points. The initial superpoint constructor of GrowSP first applies VCCS \citep{papon2013voxel} and then further merges the generated superpoints using region growing \citep{adams1994seeded} and majority voting. We denote the set of initial superpoints of point cloud $\mathbfcal{P}_i\in\mathbf{P}$ by $\mathbfcal{S}_i=\{\mathbf{s}_i^1,\ldots,\mathbf{s}_i^{M^0}\}$, where $M^0$ is the number of superpoints, which may differ between input clouds.

GrowSP utilizes a submanifold sparse convolutional neural network \citep{choy20194d} as a feature extractor, although in principle any neural network capable of extracting $K$-dimensional feature vectors from point clouds could be used. The specific architecture, ResNet16FPN, consists of a ResNet16 \citep{he2016deep} encoder and a feature pyramid network (FPN) \citep{lin2017feature} decoder. The number of extracted point-level features is set to $K=128$. The training of GrowSP has two stages: pretraining and growth. During the former stage, the feature extractor is trained for $E_{\text{pretrain}}$ epochs with the initial superpoints, and during the growth stage, the superpoint constructor begins to progressively generate larger and fewer superpoints to be used in training.

Once the pretraining has concluded, the extracted features are expected to be relatively meaningful. As such, the superpoint constructor begins merging the initial superpoints. Given a point cloud $\mathbfcal{P}_i$ with $N$ points and the corresponding superpoints $\mathbfcal{S}_i$, the constructor first generates superpoint level representations from the extracted pointwise features $\mathbfcal{F}_i\in\real^{N\times K}$ by simply computing their means within each superpoint $\mathbf{s}^j_i$:
\begin{equation}
    \mathbf{\Bar{f}}^j_i=
    \frac{1}{|\mathbf{s}^j_i|}\sum_{k\in \mathbf{s}^j_i}
    \mathbf{f}^k_i,
\end{equation}
where $\mathbf{f}^k_i\in\real^{1\times K}$ is the feature vector retrieved from $\mathbfcal{F}_i$ corresponding to the $k$th point in $\mathbf{s}_i^j$ and $|\cdot|$ denotes the cardinality of a set. The superpoint neural features are then combined with additional superpoint level descriptors, specifically coordinates $\Bar{\mathbf{p}}_i^j\in\real^{1\times 3}$, colors $\Bar{\mathbf{c}}_i^j\in\real^{1\times 3}$ (if available), and estimated surface normals $\Bar{\mathbf{n}}_i^j\in\real^{1\times 3}$:
\begin{equation}
    \Tilde{\mathbf{f}}_i^j=\Bar{\mathbf{f}}_i^j\oplus
    w_{\text{xyz}}\cdot\Bar{\mathbf{p}}_i^j\oplus
    w_{\text{rgb}}\cdot\Bar{\mathbf{c}}_i^j\oplus
    w_{\text{norm}}\cdot\Bar{\mathbf{n}}_i^j,
\end{equation}
where $w_{\text{xyz}}, w_{\text{rgb}}, w_{\text{xyz}}\in\real^+$ are scalar weights for the coordinates, colors, and surface normals respectively and $\oplus$ denotes concatenation.

The superpoint level features $\{\Tilde{\mathbf{f}}_i^0,\ldots\Tilde{\mathbf{f}}_i^{M^0}\}$ are then used for merging the $M^0$ initial superpoints into $M^j<M^0$ larger superpoints with the $k$-means clustering algorithm. The superpoint growing is conducted independently for each point cloud $\mathbfcal{P}_i$ every $\widehat{E}\in\integer^+$ epochs and the merging is conducted a total of $T$ times with the final number of superpoints $M^T<M^0$ being a user-defined hyperparameter.

At all stages of the training, each of the $H$ input clouds in the data is represented by a varying number of superpoints. This combined set of all superpoints can be viewed as a large set of primitive semantic elements, which need to be mapped into the set of actual semantic classes $C$. However, \cite{zhang2023growsp} empirically found that simply clustering the superpoints into $|C|$ categories is an excessively aggressive approach that results in poorly performing segmentation models. Consequently, the semantic primitive clustering module of GrowSP groups the superpoints into a relatively large number of semantic primitives during training. The labels of these primitives are utilized as pseudo-labels for training the feature extractor.

The semantic primitive clustering module utilizes a clustering strategy very similar to the superpoint constructor. The key difference is that since pseudo-labels are already required during the pretraining stage, the superpoint level neural embeddings $\mathbf{\Bar{f}}_i^j$ alone are not reliable enough for grouping superpoints as the extracted features are essentially random in the beginning. As such, GrowSP utilizes point feature histograms (PFH) \citep{rusu2008aligning}, which are widely used geometric point cloud descriptors based on estimated surface normals. The 10-dimensional PFHs, denoted by $\mathbf{\Ddot{f}}_i^j$, are then concatenated with other superpoint level descriptors:
\begin{equation}
    \mathbf{\Hat{f}}_i^j=\mathbf{\Bar{f}}_i^j\oplus
    w_{\text{pfh}}\cdot\mathbf{\Ddot{f}}_i^j\oplus
    w_{\text{rgb}}\cdot\mathbf{\Bar{c}}_i^j,
\end{equation}
where $w_\text{pfh}, w_\text{rgb}\in\real^+$ are scalar weights for the corresponding feature vectors. The resulting feature vectors $\{\mathbf{\Hat{f}}_i^0,\ldots,\mathbf{\Hat{f}}_i^{M^0}\}$ are subsequently clustered into $S$ semantic primitives with the $k$-means algorithm. In contrast to the superpoint constructor, all features across the entire data set $\mathbf{P}$ are clustered simultaneously.

The semantic primitive labels are stored as $S$-dimensional one-hot pseudo-labels. Furthermore, the estimated centroids of the primitives are used for constructing a linear classifier, which can be used for obtaining $S$-dimensional logits for each individual point in a given neural embedding $\mathbfcal{F}_i$. Identically to the superpoint constructor, semantic primitive clustering is conducted once every $\widehat{E}$ epochs.

During training, GrowSP first passes a batch of point clouds $\mathbfcal{B}$ through the feature extractor and then retrieves $S$-dimensional logits for each point from the semantic primitive classifier with the extracted features. Standard cross-entropy loss is computed between the logits and the previously saved pseudo-labels, and network parameters are subsequently optimized using stochastic gradient descent (SGD) with a polynomial learning rate scheduler. 

At test time, to obtain the final semantic class predictions, the $S$ centroids of the semantic primitive classifier are simply merged into $|C|$ groups with $k$-means clustering. To match the predicted classes to ground truth labels GrowSP models the task of finding an optimal mapping from the set of predicted labels to the set of ground truth labels as a linear assignment problem and subsequently solves it using the Hungarian method \citep{kuhn1955hungarian}.

\subsubsection{Proposed modifications to GrowSP} \label{section:growsp_mods}

While the generic GrowSP model can be utilized for semantic segmentation of multispectral forest point clouds to some success based on our experiments, we found that the modifications proposed in this section drastically improved segmentation accuracy.

\begin{figure*}[!ht]
\centering
\includegraphics[width=\textwidth]{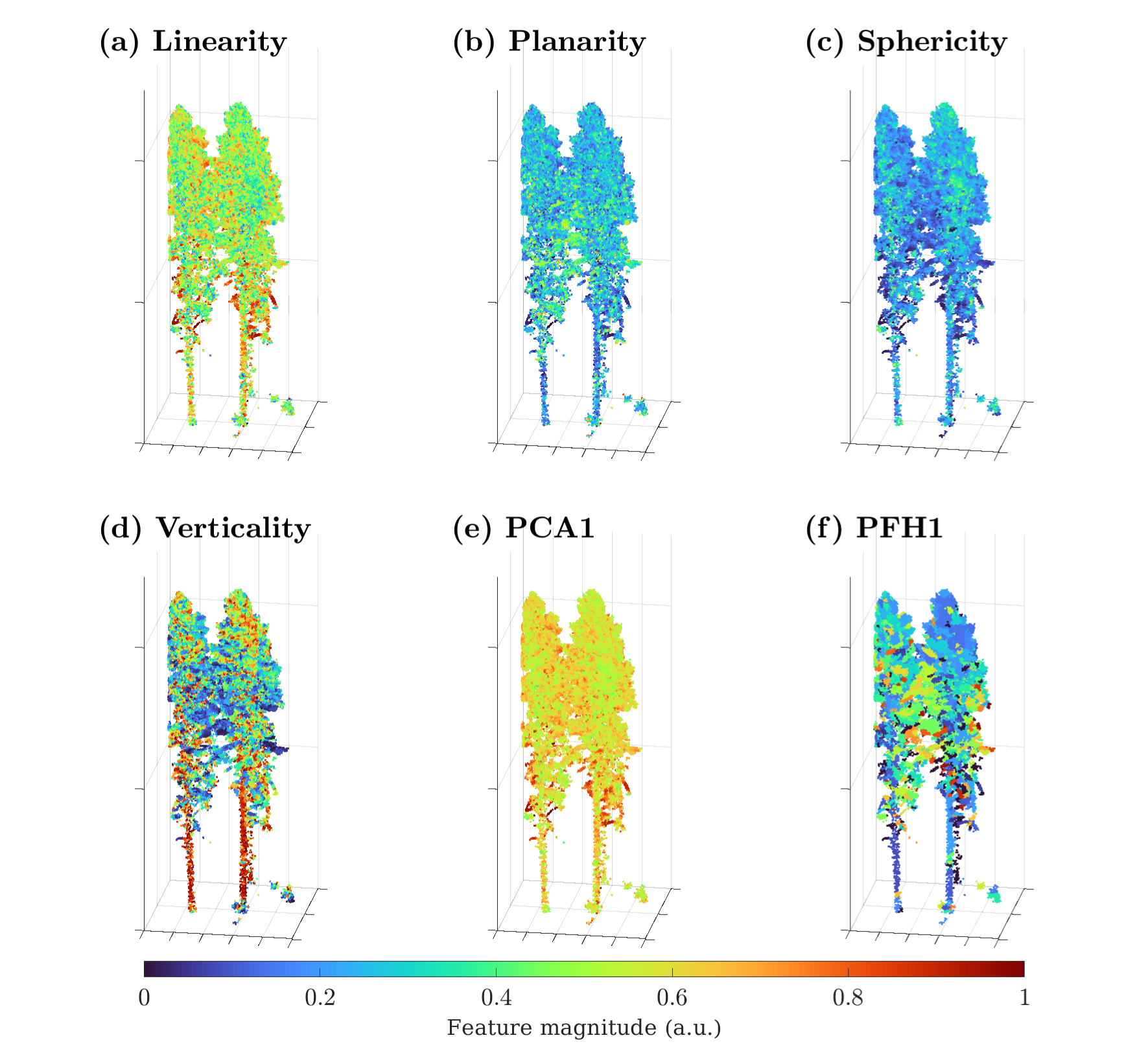}
\caption{Comparison between geometric features and point feature histogram descriptors for one cylinder from plot \#1. The geometric features have been computed for the entire plot prior to dividing it into smaller cylinders. (a) Linearity. (b) Planarity. (c) Sphericity. (d) Verticality. (e) First principal component. (d) The first point feature histogram descriptor from 10-dimensional PFHs. Note that the values have been scaled to the range [0,1] for visualization purposes.}
\label{fig:geometric_features}
\end{figure*}

\noindent\textbf{Geometric point cloud descriptors:} In order to improve the accuracy of semantic primitive clustering during the early stages of training the backbone network, GrowSP utilizes 10-dimensional point feature histograms, which are based on estimated surface normals. We found that within complex forest canopy structures the surface normals may vary highly even within small geometrically consistent regions. As such, while PFHs are suitable for segmenting point clouds with simpler geometry and solid surfaces, such as indoor scenes, they are a suboptimal choice when it comes to forest environments. Consequently, GrowSP-ForMS adopts an alternative set of geometric descriptors based on the eigenvalues of point neighborhood covariance tensors as specified by \cite{hackel2016contour}. A visual example illustrating how geometric features are more effective at distinguishing wooden components in comparison to PFHs is shown in \autoref{fig:geometric_features}.

Given a point cloud $\mathbfcal{P}\in\real^{N\times3}$, the covariance tensor of a point $p_i\in\real^3$ is defined as follows:
\begin{equation}
    \Sigma_i=\frac{1}{K}\sum_{p_k\in\mathbfcal{N}_i}(p_k-p_{\text{med}})(p_k-p_{\text{med}})^\top,
\end{equation}
where $\mathbfcal{N}_{i}=\{p_i^1,\ldots,p_i^{K}\}$ is a set comprised of the $K$ nearest neighbors of $p_i$ and $p_{\text{med}}$ is its medoid. The geometric descriptors are computed using the eigenvalues $\lambda_1\geq\lambda_2\geq\lambda_3\geq0$ and eigenvectors $\mathbf{e}_1,\mathbf{e}_2,\mathbf{e}_3$ of $\Sigma_i$. From the wide variety of descriptors that can be computed with $\Sigma_i$, we opt for linearity, planarity, sphericity, verticality and the first principal component (PCA1), which have been successfully utilized for leaf--wood separation both in the context of deep learning \citep{shen2022deep,morel2020segmentation} and unsupervised algorithms \citep{wang2020lewos,tian2022graph,wang2020unsupervised,shcherbcheva2023unsupervised}. For a given point $p_i$, the features are computed as follows:
\begin{align}
    &\text{Linearity}(p_i)=\frac{\lambda_1-\lambda_2}{\lambda_1}\\
    &\text{Planarity}(p_i)=\frac{\lambda_2-\lambda_3}{\lambda_1}\\
    &\text{Sphericity}(p_i)=\frac{\lambda_3}{\lambda_1}\\
    &\text{Verticality}(p_i)=1-\text{abs}(\hat{\mathbf{z}}^\top\mathbf{e}_3)\\
    &\text{PCA1}(p_i)=\frac{\lambda_1}{\lambda_1+\lambda_2+\lambda_3},
\end{align}
where $\hat{\mathbf{z}}=[0,0,1]$ and $\text{abs}(\cdot)$ is the absolute value \citep{hackel2016contour}.

\cite{weinmann2017geometric} found that the definition of the neighborhood $\mathbf{P}_i$ used for computing $\Sigma_i$ has a significant effect on feature relevance. When it comes to semantic segmentation, the best option is to use spherical neighborhoods parameterized by either a radius or the number of nearest neighbors. We utilize a hybrid strategy, where all points within a spherical neighborhood of radius $r_{n}$ up to the $K$ nearest neighbors are considered. Following \citet{shcherbcheva2023unsupervised}, the radius was set to $r_n=0.35$ m. With the goal of minimizing the variance of feature values caused by varying point densities, we adopt a multi-scale strategy for computing the geometric descriptors inspired by the eigenentropy-based adaptive neighborhood selection of \cite{weinmann2017classification}. Each descriptor is first computed for all $K\in\mathbfcal{K}=\{20, 50, 100, 150\}$, and the four values are subsequently compiled into a single feature vector by computing the mean:
\begin{equation}
    \overline{\text{F}}(p_i)=\frac{1}{|\mathbfcal{K}|}\sum_{K\in\mathbfcal{K}}\text{F}_K(p_i),
\end{equation}
where $\text{F}_K(\cdot)\in\real$ is some geometric descriptor computed such that the number of nearest neighbors within the spherical neighborhood has been thresholded to $K$. Although simple, based on our experiments the method appears to effectively smooth out the geometric descriptors for forest data.

\noindent \textbf{Graph-based superpoint constructor:} Ideally, the points within each initial superpoint should all represent the same ground truth class. However, this is generally never the case in practice if the initial superpoints are constructed with no supervision. Superpoints comprising points from multiple categories complicate the training process, since GrowSP generates superpoint features by averaging over point level features. As a result, GrowSP pushes the backbone network toward learning similar features for points within the same superpoint regardless of the ground truth class. Due to GrowSP's heavy reliance on the consistency of superpoints, optimizing the accuracy of the initial superpoint constructor is a crucial part of the framework. A straightforward method for assessing the accuracy of a set of superpoints is assigning each point the most common ground truth label in the corresponding superpoint and comparing the resulting set of superpoint-based labels to the ground truth.

On the other hand, since the semantic primitive clustering module clusters all superpoint-level representations, given two sets of superpoints with equal accuracy, the set with a lower number of superpoints is always preferable from the viewpoint of computational complexity. Consequently, the number of superpoints per input cloud is also a crucial metric when it comes to measuring constructor performance. For reference, with around 1,400 initial superpoints per input point cloud for our data set the clustering requires around 30 minutes in wall-clock time with our hardware (see \autoref{section:experimental_setup}).

We found that the original VCCS-based superpoint constructor performs quite poorly on ALS forest data. Depending on the hyperparameters, the constructor creates either prohibitively many superpoints or assigns a significant portion of wood points to superpoints with foliage point majority. As such, we opt to use the superpoint constructor of \cite{landrieu2018large} instead. Given a point cloud $\mathbfcal{P}$ and $d\in\integer^+$ geometric descriptors $f_i\in\real^{d}$ corresponding to each point $p_i\in\mathbfcal{P}$, the constructor first generates a 10-nearest neighbors adjacency graph $G=(V,E)$ based on Euclidean distance between the $xyz$-coordinates. The set of initial superpoints is then equivalent to the constant connected components of the solution to:
\begin{equation}\label{equation:energy_min}
    \underset{g\in\real^{d}}{\text{argmin}}\sum_{i\in V}||g_i-f_i||^2+\lambda\sum_{(i,j)\in E}w_{ij}\mathbbm{1}(g_i-g_j\neq0),
\end{equation}
where $w_\text{ij}=||p_i-p_j||_2^{-1}$ denotes the edge weight, $\lambda\in\real^+$ is the user defined regularization strength and $\mathbbm{1}(\cdot)$ is the indicator function. The superpoint constructor utilizes the $\ell_0$-cut pursuit algorithm of \cite{landrieu2017cut} for finding an approximate solution to \autoref{equation:energy_min}. Following \cite{shen2022deep} who used the same superpoint constructor for semantic segmentation of TLS forest point clouds, we opted for linearity, verticality, planarity, and sphericity as the geometric descriptors $f_i$. Based on initial experiments, the regularization strength was set to $\lambda=0.1$ and the geometric features were further weighted by multiplying them with the scalar 10.

While the superpoint constructor of \cite{landrieu2018large} alone yielded a comparatively high accuracy when applied to forest data, the number of superpoints remained prohibitively large for many input clouds. Furthermore, many of the superpoints were quite small with less than five points, and thus unlikely to contain a sufficient amount of semantic information. As such, we propose a simple algorithm for merging small superpoints to adjacent ones, which is summarized in \autoref{algorithm:superpoint_merge}.

The merging procedure utilizes an adaptive threshold $\text{PTS}_{\text{min}}\in\integer^+$ for the minimum number of points each superpoint must contain. $\text{PTS}_{\text{min}}$ is determined separately for each input cloud based on a user-defined maximum number of superpoints $\text{SP}_{\max}\in\integer^+$. All superpoints comprised of less than $\text{PTS}_{\text{min}}$ points are then classified as \emph{non-admissible} and any non-admissible superpoint with two or fewer points is further classified as \emph{singular}. Subsequently, singular superpoints are clustered into larger ones with the DBSCAN \citep{ester1996density} algorithm. Finally, each non-admissible superpoint is merged with the closest admissible superpoint based on Euclidean distance.

\begin{algorithm}
\caption{Procedure for merging the superpoints  $\mathbfcal{S}_i^\text{prelim}$ constructed with the $\ell_0$-cut pursuit algorithm such that there are at most $\text{SP}_{\text{max}}$ superpoints. The procedure $\Call{CountSuperpoints}{\mathbfcal{S}_i, x}$ returns the number of superpoints in $\mathbfcal{S}_i$ comprised of more than $x\in\integer^+$ points, while $\Call{GetNonAdmissible}{\mathbfcal{S}_i}$ and $\Call{GetSingular}{\mathbfcal{S}_i}$ obtain all non-admissible and singular superpoints in $\mathbfcal{S}_i$ respectively. The hyperparameters $\text{SP}_{\text{max}}, \epsilon$, and $\text{min}_{\text{samples}}$ were set to 2200, 0.2, and 5 respectively based on initial experiments.} \label{algorithm:superpoint_merge}
\begin{algorithmic}[1]
    \Procedure{Merge}{$\mathbfcal{S}_i^\text{prelim},\text{SP}_{\text{max}},\epsilon,\text{min}_{\text{samples}}$}
        \State $\text{PTS}_{\text{min}} \gets5$
        \State $\text{SP}_i \gets$ \Call{CountSuperpoints}{$\mathbfcal{S}_i^\text{prelim},\text{PTS}_{\text{min}}$}
        \While{$\text{SP}_i>\text{SP}_{\text{max}}$}
            \State $\text{PTS}_{\text{min}} \gets \text{PTS}_{\text{min}}+1$
            \State $\text{SP}_i \gets$ \Call{CountSuperpoints}{$\mathbfcal{S}_i^\text{prelim},\text{PTS}_{\text{min}}$}
        \EndWhile
        \State $\mathbfcal{S}_i^{\text{non-ad}} \gets \Call{GetNonAdmissible}{\mathbfcal{S}_i^\text{prelim},\text{PTS}_{\text{min}}}$
        \State $\mathbfcal{S}_i^{\text{singular}} \gets$ \Call{GetSingular}{$\mathbfcal{S}_i^{\text{non-ad}}$}
        \State $\mathbfcal{S}_i \gets \mathbfcal{S}_i^\text{prelim}\setminus\mathbfcal{S}_i^{\text{non-ad}}$
        \State $\mathbfcal{S}_i^{\text{non-ad}} \gets \mathbfcal{S}_i^\text{non-ad}\setminus\mathbfcal{S}_i^{\text{singular}}$
        \State $\mathbfcal{S}_i^{\text{singular}} \gets$ \Call{DBSCAN}{$\mathbfcal{S}_i^{\text{singular}},\epsilon,\text{min}_{\text{samples}}$}
        \State $\mathbfcal{S}_i^{\text{non-ad}} \gets \mathbfcal{S}_i^{\text{non-ad}}\oplus\mathbfcal{S}_i^{\text{singular}}$
        \ForAll{$\mathbf{s}_i^j\in\mathbfcal{S}_i^{\text{non-ad}}$}
            \State $\mathbf{s}_i^\ast \gets \text{argmin}_{\mathbf{s}_i^k\in\mathbfcal{S}_i}||\mathbf{s}_i^j-\mathbf{s}_i^k||_2$
            \State  $\mathbf{s}_i^\ast \gets \mathbf{s}_i^\ast\oplus\mathbf{s}_i^j$
        \EndFor
        \State \Return $\mathbfcal{S}_i$
    \EndProcedure
\end{algorithmic}
\end{algorithm}

\noindent \textbf{Decaying clustering weights:} As the training of GrowSP progresses, the neural features are expected to become increasingly more meaningful, while the conventional geometric features used as assistance in semantic primitive clustering should simultaneously become less important. Since the neural features are initially effectively random, geometric descriptors and color information should be weighted quite heavily early on. However, if the conventional point cloud features are weighted excessively the model fails to learn any meaningful features since the neural features have negligible influence on the semantic primitive clustering.

GrowSP-ForMS introduces an adaptive weighting strategy, where the clustering weights decay linearly as a function of the current epoch until a certain minimum weight $w_{\text{coef}}^\ast\geq0$ is reached. This way the conventional point cloud features can be weighted more heavily during early stages of training without hindering feature learning at later stages. The linear decay is achieved by multiplying the features with a weight coefficient:
\begin{equation}
    w_{\text{coef}}=\max\left(1-\frac{E}{D_{\text{coef}}},w_{\text{coef}}^\ast\right),
\end{equation}
where $E$ is the current epoch and $D_{\text{coef}}$ and $w_{\text{coef}}^\ast$ are user-defined hyperparameters, which we empirically set to 100 and 0.2 respectively. The augmented feature vectors GrowSP-ForMS uses for clustering semantic primitives are then defined as follows:
\begin{equation}
    \mathbf{\Hat{f}}_i^j=\underbrace{\mathbf{\Bar{f}}_i^j}_{\text{neural features}}\oplus
    \underbrace{w_{\text{coef}}\cdot w_{\text{geof}}\cdot\mathbf{\Ddot{f}}_i^j}_{\text{geometric descriptors}}\oplus
    \underbrace{w_{\text{coef}}\cdot w_{\text{rgb}}\cdot\mathbf{\Bar{c}}_i^j}_{\text{reflectance features}},
\end{equation}
where $\mathbf{\Ddot{f}}$ now denotes the geometric descriptors instead of the 10-dimensional PFHs, $\Bar{\mathbf{c}}_i^j\in\real^{1\times 3}$ denotes the reflectance information instead of RGB color and $w_{\text{geof}},w_{\text{rgb}}\in\real^+$ are the corresponding weights.

\noindent \textbf{Oversegmentation of predicted labels:} oversegmentation entails setting the number of clusters $k$ to a value that is higher than the number of ground truth classes $|C|$ at prediction time. The resulting labels will then contain multiple distinct classes representing foliage and wood. Assuming that the wooden components have been separated from the foliage adequately, each oversegmented class can be identified as one of the two, since a set of points comprised mainly of wood generally differs quite significantly from a set of foliage points when it comes to geometric descriptors such as linearity. We empirically observed that utilizing such an oversegmentation strategy on the semantic primitives at prediction time provides a significant improvement in segmentation accuracy. 

In practice, the oversegmentation is implemented by defining a new set of ground truth classes $C_{\text{over}}$ such that $|C_{\text{over}}|>|C|$, which is used at prediction time. Subsequently, each oversegmented class is assigned to either wood or foliage based on average linearity and user-defined linearity threshold $L_{\min}$. Conveniently, since training the backbone network of GrowSP is entirely based on the semantic primitive labels, the values of $C_{\text{over}}$ and $L_{\min}$ have no effect on the training. As such, both hyperparameters can be optimized once the model has already been trained, that is, our oversegmentation strategy does not incur any additional computational overhead during training. For our data set, we found that utilizing $|C_{\text{over}}|=14$ classes and a linearity threshold of $L_{\min}=0.55$ yielded the best performance.

\subsection{Reference leaf--wood separation models}

To assess how the performance of GrowSP-ForMS compares to existing leaf--wood separation approaches, we tested three popular fully supervised deep learning methods and two unsupervised geometry-based algorithms on our multispectral data set. This section provides a brief overview of each method.

\subsubsection{Supervised deep learning models}
We chose three popular semantic segmentation neural networks as our supervised reference methods since most DL models specifically designed for leaf--wood separation are optimized for specific data sets and rarely have openly available implementations. To ensure the reference models were adequate for leaf--wood separation, we chose architectures that have been successfully utilized for semantic segmentation of forest point clouds in some previous work.

PointNet \citep{qi2017pointnet} was chosen as the first fully supervised reference method since it has along with its successor PointNet++ served as the basis for numerous DL-based leaf--wood separation models \citep[see e.g.][]{morel2020segmentation,krisanski2021sensor}. For the second reference model, we chose RandLA-Net \citep{hu2020randlanet}, which has previously been successfully utilized for semantic segmentation of MLS forest data \citep{kaijaluoto2022semantic}. Finally, following the original GrowSP article, we utilized the feature extractor ResNet16FPN described in \autoref{section:growsp_original} as the third reference method.

\noindent \textbf{PointNet:} As a pioneering work in point cloud deep learning, PointNet was the first model to directly process unordered point clouds as input and remains quite popular to this day, especially as a performance baseline. At its core, PointNet is a multilayer perceptron (MLP), with certain additions inspired by the structural characteristics of point clouds. Notably, PointNet is both permutation and transformation invariant and captures information from local and global contexts. In our experiments, we utilized an open-source PyTorch implementation\footnote{\url{https://github.com/yanx27/Pointnet_Pointnet2_pytorch}} of PointNet.

PointNet first extracts point-level features using a collection of shared MLPs, where shared refers to applying the same MLP to each point in the input cloud. The features are then aggregated into a global feature vector through a global max pooling layer. To perform semantic segmentation, the global and pointwise features are concatenated into a single feature vector and subsequently passed as input for the segmentation network, which extracts new per-point features based on the aggregated information. This enables predicting pointwise characteristics that are reliant on both local geometry and global semantics \citep{qi2017pointnet}.

\noindent \textbf{RandLA-Net:} The RandLA-Net architecture is specifically designed for efficient yet effective semantic segmentation of large-scale 3D point clouds and therefore well suited for processing ALS data captured in a forest environment. We used an open-source PyTorch implementation of RandLA-Net\footnote{\url{https://github.com/matthiasverstraete/3d_recognizer/tree/main/randlanet}} in our experiments.

RandLA-Net achieves efficiency on large-scale data by randomly downsampling the input cloud on each neural layer. To prevent randomly discarding crucial features, RandLA-Net utilizes a novel local feature aggregation (LFA) module, which is applied in tandem with the downsampling. The LFA module preserves prominent features by aggregating information within point neighborhoods. Each module contains multiple blocks composed of \emph{local spatial encoding} (LocSE) units and \emph{attentive pooling} layers. A LocSE unit generates a relative point position encoding for each input point using its $k$ nearest neighbors, while an attentive pooling layer aggregates the set of neighboring point features by automatically learning the most important local features using a shared MLP \citep{hu2020randlanet}.

\subsubsection{Unsupervised leaf--wood separation algorithms}

For the unsupervised reference methods, we chose two recent point cloud geometry-based leaf--wood separation algorithms LeWoS \citep{wang2020lewos} and GBS \citep{tian2022graph}. The algorithms were chosen because their official implementations were publicly available and both have reportedly performed well for a wide variety of tree species.

\noindent \textbf{LeWoS:} LeWoS is an unsupervised leaf--wood separation algorithm based on point cloud connectivity and geometric descriptors. The parameter tuning is straightforward, as LeWoS requires only one user-defined parameter, i.e., the similarity threshold $N_{\text{z\_thres}}$ and is robust toward the selection. We used the official MATLAB implementation\footnote{\url{https://github.com/dwang520/LeWoS}} in all of our experiments.

The semantic segmentation process of LeWoS consists of three fundamental steps: recursive graph segmentation, class probability estimation, and class regularization. The recursive graph segmentation begins by constructing an initial graph with an edge between each point and its ten nearest neighbors. Subsequently, the edges are pruned based on feature similarity and distance after which the pruned graph is segmented into clusters by finding its connected components. The graph construction and segmentation is then applied recursively on the resulting clusters until no more clusters can be further segmented or a maximum of ten iterations. Finally, LeWoS splits any remaining large branch clusters by applying the small cover set-based method of \cite{raumonen2013fast}.

Subsequent to performing the recursive point cloud segmentation, LeWoS estimates a class probability for each cluster based on a predefined set of linearity and size thresholds. The estimated probabilities form a \emph{soft labeling} set $S$, where each point is represented by two probability values denoting the wood and foliage classes respectively. Due to factors such as occlusions and abnormal growth angles, directly classifying each point to the class with the highest probability is often suboptimal. Consequently, in an effort to achieve a more spatially consistent labeling set, LeWoS applies a class regularization on $S$:
\begin{equation}\label{equation:lewos_reg}
    S^\prime=\underset{O\in\Omega^V}{\text{argmin}}\left(\Phi(S,S^\prime)+\gamma\Psi(S^\prime)\right),
\end{equation}
where $\Phi(\cdot)$ is a linear fidelity term, $\Psi(\cdot)$ is a regularizer that favours spatially smooth solutions, $\gamma>0$ is the user defined regularization strength, and $\Omega$ is the solution search space. \autoref{equation:lewos_reg} is then solved using the $\alpha$-expansion algorithm, which yields the set of improved and spatially smoothed leaf--wood labels $S^\prime$.

\noindent \textbf{GBS:} Much like LeWoS, GBS performs leaf--wood separation based on point cloud connectivity and geometric descriptors, although the strategy is quite different. The four main steps of GBS are graph construction, multi-scale segmentation, initial wood point extraction, and region growing. In our experiments, we utilized the official implementation of GBS, which the authors provide as a Python package \citep{tian2022graphcode}. 

As a first step, GBS constructs a graph $G=(V,E)$ from the input point cloud. However, instead of utilizing a simple $k$-nearest neighbors approach, GBS adopts a novel graph construction algorithm from the open-source Python library TLSeparation \citep{vicari2017tlseparation}, which constructs a connected graph with fewer edges and is more robust to occlusions. Subsequently, GBS applies Dijkstra's shortest path algorithm to find the shortest paths from all points to the root point, that is, the lowest point in the cloud. The graph is then pruned by removing edges where either the distance or angle between subsequent points in the shortest path is above a certain threshold.

Having constructed the graph, GBS segments the input cloud into equal interval layer bins based on the shortest path length and subsequently clusters each bin by finding the connected components of the pruned graph. The segmentation is performed with multiple distinct interval sizes, each of which corresponds to a separate single-scale segmentation result. The set of intervals $I$ is a user-defined parameter.

Following the multi-scale segmentation, initial wood points are extracted separately from each single-scale segmentation result based on the size, cylindricity, and linearity of the connected components. The initial wood point labels are then corrected through a heuristic procedure based on the assumed structure of branches. To maximize the benefits of the correction, it is generally applied a total of three times. As a last step, GBS generates the final wood labels by applying a region growing algorithm on the pruned graph, where the initial wood points from the previous step are used as seed points.

\subsection{Experimental setup} \label{section:experimental_setup}

GrowSP-ForMS and all the reference models were evaluated on our multispectral data set to compare their performance in the task of leaf--wood separation. The annotated training split was utilized for both training the fully supervised DL models and optimizing the hyperparameters of unsupervised algorithms. Due to the limited amount of labeled data, we opted not to have a separate validation set, which is typically used to determine when to finish training a DL model. Consequently, the training was simply halted when the training loss had seemingly stabilized.

All experiments were carried out on a system with Intel\textsuperscript{{\textregistered}} Xeon\textsuperscript{{\textregistered}} w5-3425 CPU and a single NVIDIA\textsuperscript{{\textregistered}} RTX\textsuperscript{\texttrademark} A6000 GPU with 48 GB of GDDR6 memory. Further specifications of the system and software are listed in \autoref{table:hardware_software}.

\begin{table}[!ht]
    \centering
    \caption{Computing hardware and software specifications of the system used in the experiments.} \bigskip
    \footnotesize
    \begin{tabular}{ll}
        \toprule
        \textbf{Device} & \textbf{Specifications} \\ \midrule \midrule
        CPU & Intel\textsuperscript{{\textregistered}} Xeon\textsuperscript{{\textregistered}} w5-3425 \\
        GPU & NVIDIA\textsuperscript{{\textregistered}} RTX\textsuperscript{\texttrademark} A6000 48 GB GDDR6 \\
        Memory & $8\times64$ GB DDR5 4800 MHz \\ \toprule
        \textbf{Software} & \\ \midrule \midrule
        Ubuntu (OS) & 22.04.3 LTS \\
        GPU Driver & 535.154.05 \\
        CUDA & 11.3.1 \\
        Python & 3.8.12 \\
        Conda & 23.5.2 \\
        pip & 21.2.4 \\
        PyTorch & 1.10.2 \\
        Torchvision & 0.11.3 \\
        Minkowski Engine & 0.5.4 \\
        MATLAB & 9.14.0.2337262 (R2023a) Update 5 \\
        \bottomrule
    \end{tabular}
    \label{table:hardware_software}
\end{table}

\subsubsection{Evaluation metrics}

To objectively compare the segmentation performance of the different semantic segmentation approaches, we adopted the evaluation metrics most frequently used in the literature concerning point cloud semantic segmentation: overall accuracy (oAcc), mean accuracy (mAcc), intersection over union (IoU), and mean intersection over union (mIoU).

Overall accuracy measures the proportion of points that have been classified correctly across the entire data set, whereas mean accuracy describes the average over the class-wise accuracies. The two metrics are defined as follows:
\begin{align}
    &\text{oAcc}=\frac{1}{N}\sum_{i=1}^N\indicatorpred{y}{i}\\
    &\text{mAcc}=\frac{1}{|C|}\sum_{c\in C}\left(\frac{1}{N_c}\sum_{i=1}^N\indicator{y}{i}{c}{}\cdot\indicatorpred{y}{i}\right),
\end{align}
where $y_i$ and $\hat{y}_i$ denote the ground truth label and predicted label of the $i$th data point respectively. Furthermore, $N_c$ denotes the number of points in class $c\in C$.

The intersection over union of class $c$, often also referred to as the Jaccard index, is defined as follows:
\small{
\begin{equation}
    \text{IoU}_c=\frac{
        \sum_{i=1}^N
            \indicator{y}{i}{c}{}\cdot\indicatorpred{y}{i}
    }{
        \sum_{i=1}^N
            \indicator{y}{i}{c}{} +
            \indicator{\hat{y}}{i}{c}{} -
            \indicator{y}{i}{c}{}\cdot\indicatorpred{y}{i}
    }.
\end{equation}
}\normalsize
The mean intersection over union can then be determined by computing the mean of the class-wise IoUs:
\begin{equation}
    \text{mIoU}=\frac{1}{|C|}\sum_{c\in C}\text{IoU}_c.
\end{equation}

\subsubsection{Experimental setup of GrowSP-ForMS} \label{section:training_growsp}

When training GrowSP-ForMS, in addition to the labeled training split, we utilized the unlabeled data from the remaining 18 test sites. This was done to improve model performance, as DL models trained on larger data sets often generalize better. Since GrowSP-ForMS is fully unsupervised, it could be trained with unlabeled data, unlike the supervised reference methods.

The configuration used for training GrowSP-ForMS was largely the same as that used by the original GrowSP \citep{zhang2023growsp}. The voxel size of the SCNN feature extractor was set to 0.05 m and an SGD optimizer with an initial learning rate of 0.1, momentum of 0.9 and a polynomial learning rate scheduler was used for minimizing the standard cross-entropy loss. We ran experiments for a wide variety of input feature combinations and ultimately chose the combination that yielded the highest training set mIoU as our final model. Perhaps interestingly, by far the best-performing setup only utilized reflectance information from scanner 1 as input feature, while simultaneously using all three reflectance channels for semantic primitive clustering and superpoint merging. We refer to this setup as the \emph{full} GrowSP-ForMS model. Further information on the experimental setup of GrowSP-ForMS can be found in \appendixref{appendix:nn_details}.

\subsubsection{Experimental setup of supervised deep learning models} \label{section:training_nn}

To achieve comparable results with all fully supervised baselines, the training process was standardized. The cylinders of radius $r_c=4.2$ m were used as input data with the $xyz$-coordinates and reflectance values from all three scanners as input features. Similarly to GrowSP-ForMS, the missing reflectance values were populated using the strategy proposed in \autoref{section:data_preprocessing}, since preliminary testing indicated such data provided slightly more accurate segmentation results in comparison to the original data with missing values. Finally, the same optimization back end as well as data augmentation techniques, specifically random rotation, scaling and translation, were adopted for all models.

To alleviate the effects of the considerable class imbalance in our MS data set, we chose to minimize focal loss \citep{lin2020focal} with $\gamma=2$ when training ResNet16FPN and RandLA-Net. For PointNet, we opted for the standard cross-entropy loss instead, as it provided better performance. Following the training framework of \cite{kaijaluoto2022semantic} for supervised semantic segmentation of MLS forest point clouds, we used an Adam optimizer \citep{kingma2014adam} and the cyclical learning rate scheduler of \cite{smith2017cyclical} for training the supervised models. For additional details of the training process, see \appendixref{appendix:nn_details}.

\subsubsection{Experimental setup of reference unsupervised algorithms}

Similarly to many other leaf--wood separation algorithms \citep[see e.g.][]{cote2009structural,tao2015geometric}, GBS requires the input clouds to represent individual trees. As such, we constructed an alternative data split by assigning each individual tree to the training or test set based on the cylinder that contained a majority of points. All points that were not a part of any individual tree were simply discarded.

In contrast to GBS, LeWoS is fully applicable to plot-level forest point clouds, although it was originally developed and tested on a data set composed of individual tropical trees \citep{wang2020lewos}. Consequently, we opted to evaluate the performance of LeWoS separately for both individual trees and the full cylinders in an effort to provide a more accurate comparison with GBS and GrowSP-ForMS.

Although the authors of LeWoS and GBS presented recommended hyperparameter values for their algorithms, both were specifically optimized for TLS data, which significantly differs from our ALS data set in terms of density and geometric consistency. As such, the hyperparameter values had to be re-optimized to provide a fair assessment of segmentation performance. Details of the parameter optimization process are provided in \appendixref{appendix:hyperparameter_optimization}.

\section{Results} \label{section:results}

In this section we first assess the performance of our graph-based superpoint constructor, followed by a quantitative and qualitative comparison between GrowSP-ForMS and the reference leaf--wood separation methods. Finally, we perform two ablation studies that examine the effectiveness of our proposed changes to GrowSP and the benefits of utilizing multispectral data.

\subsection{Graph-based superpoint constructor}

\begin{figure*}[!ht]
\centering
\includegraphics[width=\textwidth]{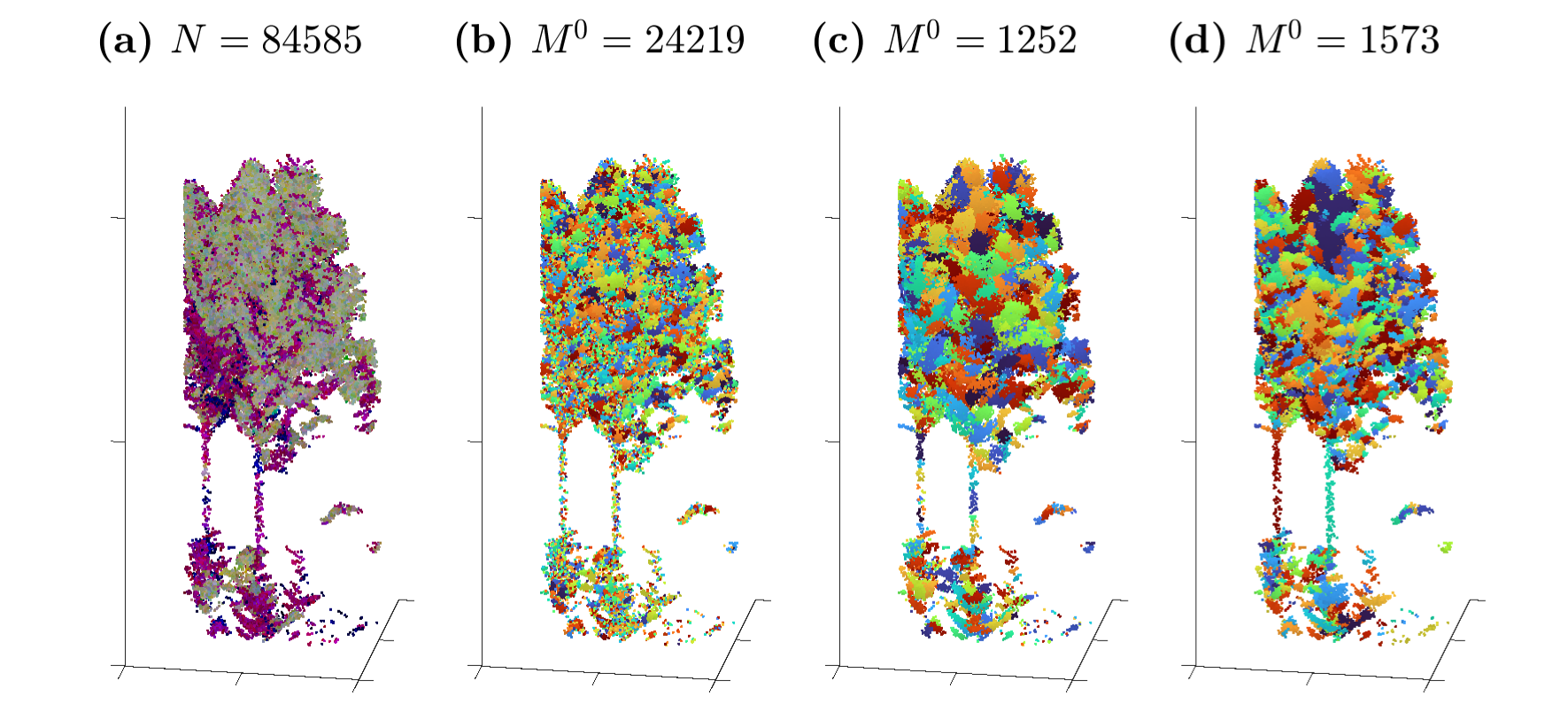}
\caption{Visual comparison of the different superpoint constructors for a sample cylinder from plot \#2 of our multispectral data set. (a) Input point cloud. $N$ denotes the number of points. (b) Superpoints generated by the VCCS-based superpoint constructor with a voxel size of 0.05 m. (c) Superpoints generated by the VCCS-based superpoint constructor with a voxel size of 0.1 m. (d) Superpoints generated by the graph-based superpoint constructor. In figures (b)--(d), $M^0$ denotes the number of superpoints.}
\label{fig:sp_comp_vis}
\end{figure*}

A comparison between our graph-based superpoint constructor (\autoref{section:growsp_mods}) and the VCCS-based constructor used in the original GrowSP architecture is shown in \autoref{table:sp_constructor_comparison}. Similarly, \autoref{fig:sp_comp_vis} shows a visual comparison of the superpoints constructed by each method.

\begin{table*}[!ht]
    \centering
    \caption{Comparison of the default VCCS-based superpoint constructor used in the original GrowSP model and the proposed graph-based constructor (\autoref{section:growsp_mods}). Accuracy metrics were computed by assigning each point the majority ground truth label within the respective superpoint. $\overline{M^0}$ denotes the mean number of superpoints across all input point clouds. For the VCCS-based superpoint constructors, the number in brackets denotes the voxel size utilized for discretizing the point cloud in meters.} \bigskip
    \small
    \begin{tabular}{cccccccc}
        \toprule
        \textbf{Constructor} & \textbf{Data} & \textbf{$\overline{M^0}$} & \textbf{OA (\%)} & \textbf{mAcc (\%)} & \textbf{mIoU (\%)} & \textbf{Foliage (\%)} & \textbf{Wood (\%)} \\ \midrule \midrule
        VCCS (0.05) & Training & 18640 & 98.6 & 96.2 & 85.4 & 98.5 & 72.2 \\
        & Test & 22298 & 98.6 & 95.6 & 86.5 & 98.6 & 74.5 \\
        & All & 19360 & 98.6 & 96.0 & 85.7 & 98.5 & 72.8 \\ \midrule
        VCCS (0.1) & Training & 1441 & 97.5 & 91.3 & 76.0 & 97.5 & 54.5 \\
        & Test & 1182 & 97.3 & 91.5 & 74.9 & 97.3 & 52.6 \\
        & All & 1390 & 97.5 & 91.3 & 75.7 & 97.4 & 54.0 \\ \midrule
        Graph-based & Training & 1345 & 98.1 & 92.4 & 81.3 & 98.0 & 64.6 \\
        & Test & 1417 & 98.0 & 92.5 & 81.1 & 97.9 & 64.2 \\
        & All & 1360 & 98.0 & 92.4 & 81.2 & 98.0 & 64.5 \\ \bottomrule
    \end{tabular}
    \label{table:sp_constructor_comparison}
\end{table*}

While utilizing the VCCS-based superpoint constructor with the suggested point cloud discretizing voxel size of 0.05 meters produced a great accuracy, the nearly 20,000 superpoints per input cloud on average was prohibitively high for training GrowSP. Doubling the voxel size to 0.1 meters reduced the average number of initial superpoints to 1,400, which our experiments showed is close to the upper bound for a reasonable number of superpoints. However, as can be seen from \autoref{table:sp_constructor_comparison}, this also significantly decreased the accuracy, yielding an mIoU of 75.7\% with a relatively low IoU of 54.0\% for the wood class. By contrast, the graph-based superpoint constructor also produced an average of 1,400 superpoints with a considerably higher mIoU of 81.2\% (+5.5 percentage points (pp)) and wood point IoU of 64.5\% (+10.5 pp). The effects of the more accurate initial superpoints on model performance are assessed in \autoref{section:ablation_changes}.

\subsection{Performance of GrowSP-ForMS and comparison to reference models} \label{section:model_comparison}

Quantitative performance metrics of the full GrowSP-ForMS model and  all reference methods are listed in \autoref{table:model_comparison_test}. It should be noted that performance metrics of models that require individual trees as input (denoted by the asterisk) are not strictly comparable to models that can process entire forest scenes since the train and test data sets are slightly different. For class-wise segmentation accuracies and train split performance metrics, the interested reader is referred to \appendixref{appendix:quantitative}.

\begin{table*}[!ht]
    \centering
    \caption{Quantitative results of different approaches on the \textbf{test split} of the proposed multispectral data set. The best accuracy metrics in each category of supervision have been highlighted. The asterisk ($^\ast$) denotes results when the inputs are individual trees, which are not strictly comparable to others due to the different data format.} \bigskip
    \small{
    \begin{tabular}{lccccc}
        \toprule
        \textbf{Model} & \textbf{OA (\%)} & \textbf{mAcc (\%)} & \textbf{mIoU (\%)} & \textbf{Foliage (\%)} & \textbf{Wood (\%)} \\ \midrule \midrule
        \textbf{Supervised methods} &&&&& \\ \midrule
        PointNet \citep{qi2017pointnet} & 95.6 & 79.8 & 63.5 & 95.5 & 31.5 \\
        RandLA-Net \citep{hu2020randlanet} & 97.0 & 88.2 & 74.1 & 96.9 & 51.3 \\
        ResNet16FPN \citep{choy20194d} & \textbf{97.5} & \textbf{90.1} & \textbf{78.5} & \textbf{97.4} & \textbf{59.5} \\ \midrule \midrule
        \textbf{Unsupervised methods} &&&&& \\ \midrule
        LeWoS \citep{wang2020lewos} & 93.3 & 58.3 & 50.7 & 93.3 & 8.0 \\
        LeWoS$^\ast$ \citep{wang2020lewos} & 93.7 & 63.4 & 51.7 & 93.7 & 9.7 \\
        GBS$^\ast$ \citep{tian2022graph} & 94.4 & 72.4 & 60.2 & 94.3 & 26.1 \\
        \textbf{GrowSP-ForMS (Ours)} & \textbf{96.4} & \textbf{84.3} & \textbf{69.6} & \textbf{96.2} & \textbf{42.9} \\ \bottomrule
    \end{tabular}}
    \label{table:model_comparison_test}
\end{table*}

\begin{figure*}[!ht]
\centering
\includegraphics[width=0.75\textwidth]{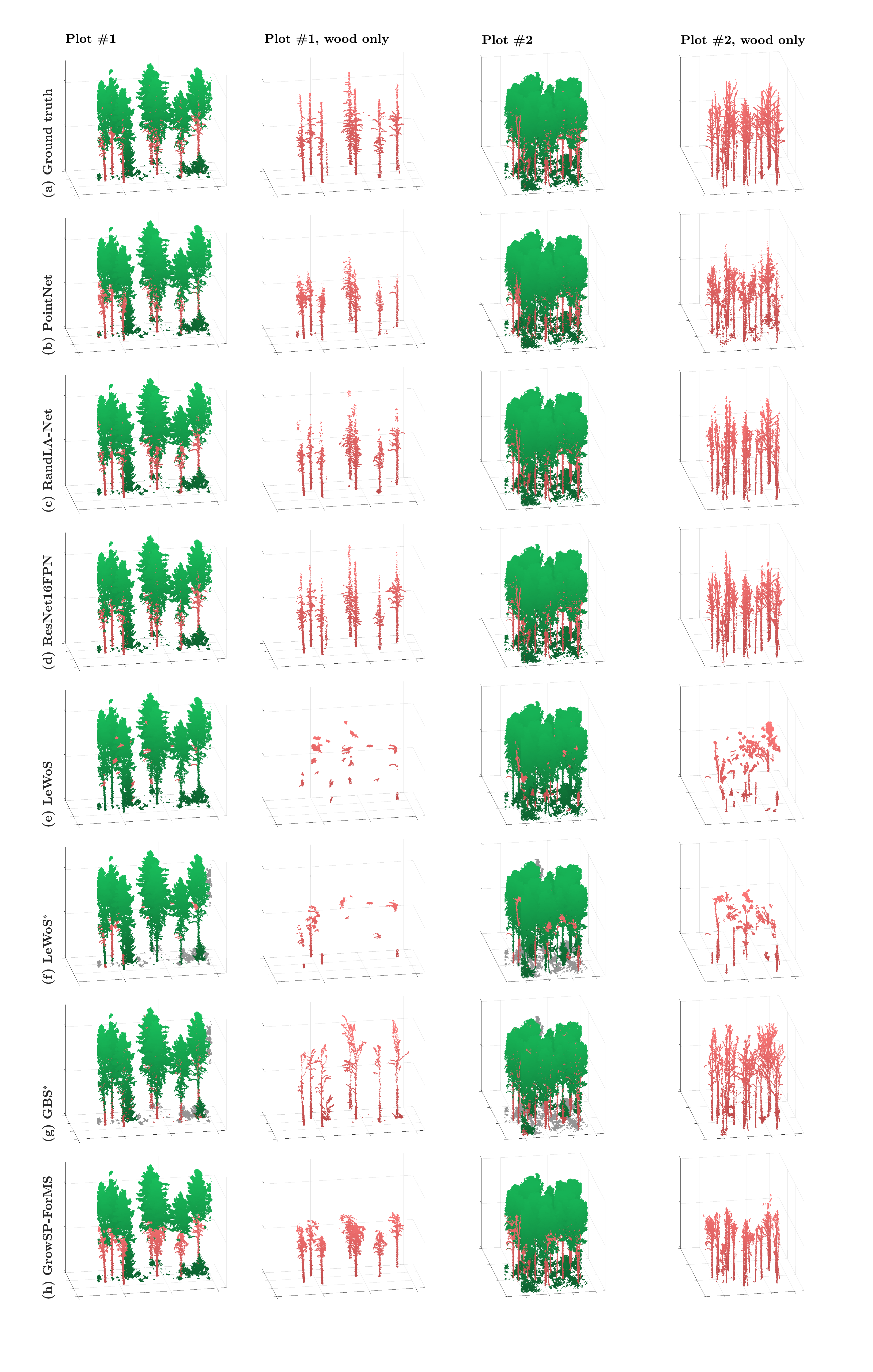}
\caption{Qualitative results of GrowSP-ForMS and our reference leaf--wood separation models for a subsection of the \textbf{test split} of plots \#1 and \#2. The first and third columns show the full segmentation results, while the second and fourth columns contain only the points classified as wood. The asterisk ($^\ast$) denotes results when the inputs are individual trees. Points colored in gray are not part of any individual tree and were thus not classified by the algorithm.}
\label{fig:performance_comparison_all}
\end{figure*}

Based on the quantitative accuracy metrics, LeWoS performed quite poorly on our multispectral ALS data set, achieving a test set mIoU of around 50\%. Both input types, i.e. individual trees and entire forest scenes resulted in similar performance. GBS performed significantly better, reaching an mIoU of 60.2\% on the test set. Nevertheless, GrowSP-ForMS outperformed both unsupervised reference methods by a considerable margin in terms of both test set mAcc (+11.9 pp) and mIoU (+9.4 pp). Consequently, GrowSP-ForMS can be considered a new state-of-the-art method in the task of unsupervised leaf--wood separation of high-density multispectral ALS point clouds.

Compared to the supervised baselines, the performance of GrowSP-ForMS was similar to PointNet. On the other hand, both RandLA-Net and ResNet16FPN clearly outperformed GrowSP-ForMS in terms of test split mIoU (-4.5 pp and -8.9 pp respectively). These performance differences are consistent with the metrics reported on multiple benchmark data sets in the original GrowSP paper \citep{zhang2023growsp}.

\autoref{fig:performance_comparison_all} shows a visual comparison of segmentation results for GrowSP-ForMS and all reference leaf--wood separation methods. The qualitative results of LeWoS were consistent with the quantitative metrics. That is, only highly linear sections of the trunk and branches with minimal occlusions were successfully detected as wood. Conversely, while the results of GBS appeared qualitatively excellent at first glance, this was not reflected by the quantitative metrics. Careful examination of the predictions revealed that GBS had a tendency to detect non-existent branches within the tree crown, while simultaneously often failing to detect clearly visible branches in lower sections of the tree. Additionally, only the very center of each trunk was detected consistently and the remaining trunk points were generally misclassified as foliage.

GrowSP-ForMS provided segmentation results that were visually more consistent than those of LeWoS and GBS. The model usually correctly detected a majority of the trunk as well as lower branches, although predictions were often overly smooth, especially within branches, as nearby foliage points were erroneously classified as wood. Another notable error source were the upper sections of the trunk, which the model consistently failed to detect. However, considering that even the best supervised baselines RandLA-Net and ResNet16FPN seemingly struggled with the same sections, we conclude that the upper trunk is overall rather difficult to classify correctly.

Much like the quantitative performance metrics, the segmentation results of PointNet were qualitatively very similar to GrowSP-ForMS. Both models yielded overly smooth predictions within branches and seldom succeeded in detecting the upper sections of the trunk. On the other hand, RandLA-Net and ResNet16FPN outperformed GrowSP-ForMS quite clearly, as was the case with the quantitative performance. Overall, the results of the two models were comparable to each other, and appeared to be quite close to the ground truth, although ResNet16FPN generally detected slightly more branches and upper proportions of the trunks. 

Additional visual comparisons of semantic segmentation results can be found in \appendixref{appendix:visual}.

\subsection{Ablation study on the proposed changes to GrowSP} \label{section:ablation_changes}

In order to accurately assess the usefulness of our proposed changes to GrowSP, starting with the original GrowSP architecture, we iteratively augmented the model with our improvements presented in \autoref{section:growsp_mods} one by one. The performance metrics of the ablated models are listed in \autoref{table:ablation_params}. While training the ablated models, we utilized the same input feature combination that was used for the full GrowSP-ForMS model. A visual comparison of the segmentation results can be found in \appendixref{appendix:ablated_visualization}.

The quantitative performance metrics suggest that each of our proposed changes to GrowSP individually improves the segmentation accuracy. This implies that all of our proposed improvements can be beneficial in the context of leaf--wood separation from multispectral ALS point clouds. Oversegmentation of predicted labels and replacing PFHs with geometric features provided the most significant improvements in model performance (+17.0 pp and +5.2 pp in mIoU respectively). The graph-based superpoints also yielded a slight improvement in segmentation accuracy (+2.6 pp in mIoU), especially for the wood class (+5.5 pp in IoU). Interestingly, the performance improvement provided by decaying clustering weights was much more significant when geometric features were used instead of PFHs (+9.8 pp and +2.7 pp in mIoU respectively).

\begin{table*}[!ht]
    \centering
    \caption{Ablation study on our proposed modifications to the original GrowSP architecture. Note that in experiments without the new graph-based superpoint constructor, we used the original VCCS-based constructor with a voxel size of 0.1 m. Similarly, 10-dimensional point feature histograms were utilized in all experiments without geometric features. Setup 1 corresponds to the original GrowSP model. The accuracies are reported for the \textbf{test split} of our multispectral data set. The best accuracy metrics have been highlighted.}
    \bigskip
    \tiny
    \begin{tabular}{P{0.7cm}P{2.6cm}P{2cm}P{1.6cm}P{1.8cm}ccccc}
        \toprule
        \textbf{Setup} & \textbf{Over\-seg\-men\-ta\-ti\-on of predicted labels} & \textbf{Graph-based superpoints} & \textbf{Geometric features} & \textbf{Decaying clustering weights} & \textbf{oAcc (\%)} & \textbf{mAcc (\%)} & \textbf{mIoU (\%)} & \textbf{Foliage (\%)} & \textbf{Wood (\%)} \\ \midrule \midrule
        \textbf{1} &&&&& 73.1 & 52.4 & 40.2 & 72.4 & 7.9 \\
        \textbf{2} & $\checkmark$ &&&& 93.3 & 66.2 & 57.2 & 93.2 & 21.1 \\
        \textbf{3} & $\checkmark$ & $\checkmark$ &&& 93.1 & 67.4 & 59.8 & 92.9 & 26.6 \\
        \textbf{4} & $\checkmark$ & $\checkmark$ & $\checkmark$ && 95.7 & 80.5 & 65.0 & 95.6 & 34.5 \\
        \textbf{5} & $\checkmark$ & $\checkmark$ && $\checkmark$ & 95.6 & 81.4 & 62.5 & 95.5 & 29.4 \\
        \textbf{6} & $\checkmark$ & $\checkmark$ & $\checkmark$ & $\checkmark$ & \textbf{96.4} & \textbf{84.3} & \textbf{69.6} & \textbf{96.2} & \textbf{42.9} \\
        \bottomrule
    \end{tabular}
    \label{table:ablation_params}
\end{table*}

\subsection{Ablation study of the effect of multispectral information} \label{section:ablation_ms}

To assess the benefits of multispectral data, we trained the GrowSP-ForMS model for each reflectance channel and all two- and three-channel combinations. In contrast to the experiments described in previous sections, where only the first reflectance channel was used as an input feature (see \autoref{section:training_growsp}), all listed channels were used both as input features and for clustering and merging of superpoints. The accuracy metrics of the ablated models are listed in \autoref{table:ablation_scanners_test}. The corresponding metrics on the train split can be found in \appendixref{appendix:quantitative}.

The results of this ablation study were not as straightforward, as the segmentation accuracy did not consistently increase with each added reflectance channel. The three best-performing scanner combinations in terms of test set mIoU were 1 \& 3, 3 and 2 \& 3. However, the former two models were quite unstable as their training set mIoUs were notably worse (-7.4 pp and -10.0 pp respectively). On the other hand, the model using the combination 2 \& 3 did not exhibit such behavior, which was also the case for the full GrowSP-ForMS model. 

The results suggest that utilizing multispectral data for leaf--wood separation can be challenging in terms of model stability. Nevertheless, MS data is seemingly beneficial for semantic segmentation accuracy, as evidenced by the full GrowSP-ForMS configuration which outperformed the best monospectral model (scanner 3) by a considerable margin (+5.8 pp in test set mIoU).

\begin{table*}[!ht]
    \centering
    \caption{Ablation study on the effects of using mono- and multispectral data. The accuracies are reported for the \textbf{test split} of our multispectral data set. The best accuracy metrics have been highlighted.}
    \bigskip
    \footnotesize
    \begin{tabular}{cccccccc}
        \toprule
        \textbf{Scanner 1} & \textbf{Scanner 2} & \textbf{Scanner 3} & \textbf{oAcc (\%)} & \textbf{mAcc (\%)} & \textbf{mIoU (\%)} & \textbf{Foliage (\%)} & \textbf{Wood (\%)} \\ \midrule \midrule
        $\checkmark$ &&& 93.7 & 69.7 & 61.9 & 93.5 & 30.2 \\
        & $\checkmark$ && 94.6 & 72.9 & 62.7 & 94.5 & 31.0 \\
        && $\checkmark$ & 94.2 & 71.6 & 63.8 & 94.0 & 33.6 \\
        $\checkmark$ & $\checkmark$ && 92.3 & 67.0 & 60.6 & 92.1 & 29.2 \\
        $\checkmark$ && $\checkmark$ & \textbf{96.3} & \textbf{86.9} & \textbf{66.9} & \textbf{96.2} & \textbf{37.5} \\
        & $\checkmark$ & $\checkmark$ & 93.5 & 69.9 & 63.1 & 93.3 & 33.0 \\
        $\checkmark$ & $\checkmark$ & $\checkmark$ & 93.5 & 69.5 & 62.5 & 93.2 & 31.7 \\
        \bottomrule
    \end{tabular}
    \label{table:ablation_scanners_test}
\end{table*}

\section{Discussion} \label{section:discussion}

Based on the qualitative segmentation results, most models, GrowSP-ForMS included, performed notably worse on test plot \#2. We conjecture that this is a result of two different factors: firstly, test plot \#2 has denser vegetation and is therefore more challenging to segment. Secondly, plot \#1 is predominantly pine, while plot \#2 is mostly composed of spruce and birches, the latter of which are likely more difficult to semantically segment due to their inherently more complicated structure. This hypothesis is supported by the work of \cite{xiang2024automated}, who observed a notably lower panoptic segmentation accuracy for plots with more complex branch structures. The imbalanced composition of the unlabeled portion of the training data set provides an alternative explanation for our observation: due to the lack of data from forest plots dominated by birch and other deciduous species, a significant majority of the unlabeled data used for training GrowSP-ForMS consists of coniferous dominated forests. Consequently, the model may have learned to distinguish wooden components of coniferous trees more effectively.

Effectively all prior existing unsupervised leaf--wood separation algorithms are designed primarily for TLS data, although some, such as LeWoS, explicitly state they also function with ALS data. Consequently, unsatisfactory performance on sparser ALS data is to be expected. In fact, \cite{tian2022graph} observed that the segmentation accuracy of both LeWoS and GBS quickly began to deteriorate when the original input cloud was subsampled. A similar phenomenon was also observed by \cite{vicari2019leaf} for their leaf--wood separation algorithm. Nevertheless, in the absence of other approaches, the existing unsupervised algorithms represent the state-of-the-art when it comes to unsupervised leaf--wood separation from ALS data, making GrowSP-ForMS the new state-of-the-art approach. However, no conclusions should be drawn about the performance of GrowSP-ForMS with any other type of data, including TLS point clouds. Future work should focus on generalizing GrowSP-ForMS to data from other LiDAR systems, including TLS and monospectral point clouds, which would facilitate a more objective performance comparison between unsupervised algorithms and allow using forest data semantic segmentation benchmarks such as FOR-Instance \citep{puliti2023forinstance}.

Although the results of our study are promising, fully supervised state-of-the-art leaf--wood separation methods outperform GrowSP-ForMS by a considerable margin. Fortunately, there are numerous prospects for future research that could help bridge the gap. Firstly, we opted to use the same backbone network for feature extraction as the original GrowSP. However, ResNet16FPN is not necessarily the optimal choice for forest data. A network designed specifically for leaf--wood separation, such as LWSNet \citep{jiang2023lwsnet} or MDC-Net \citep{dai2023mdcnet}, or a more recent transformer-based architecture such as PTv3 \citep{wu2024point} may be better suited for feature extraction. On the other hand, utilizing an SE(3) equivariant neural network \citep[see e.g.][]{fuchs2020se3,du2022se3} for feature extraction could yield more stable features in an unsupervised setting and by extension improve model performance. Secondly, the addition of geometric input features, such as linearity, planarity and sphericity (see \autoref{section:growsp_mods}), should be experimented with, as multiple studies have found they improve the segmentation accuracy of DL-based leaf--wood separation \citep{jiang2023lwsnet,dai2023mdcnet,xiang2024automated}. Finally, the addition of an unsupervised contrastive pretraining stage similar to, for example, \cite{xie2020pointcontrast} and \cite{hou2021exploring} should be considered in future work. A pretrained feature extractor could provide more descriptive features already in the early stages of training, eliminating the need for geometric point cloud features entirely and making the model more data-driven and less dependent on heuristic geometric descriptors.

\section{Conclusions} \label{section:conclusions}

In this study, we proposed the first unsupervised deep learning approach for semantic segmentation of multispectral ALS forest point clouds. Our proposed method, GrowSP-ForMS, was adapted from the original unsupervised GrowSP architecture by introducing modifications designed to improve its performance in the task of leaf--wood separation. The proposed improvements included a new graph-based superpoint constructor, utilizing alternative geometric descriptors, weight decay during semantic primitive clustering, and oversegmentation of predictions combined with linearity thresholding for detecting wooden components.

The performance of GrowSP-ForMS was compared to three supervised reference methods PointNet, RandLA-Net, and ResNet16FPN, and two unsupervised algorithms LeWoS and GBS, all of which were applied to the same data set. GrowSP-ForMS achieved a test set mIoU of 69.6\%, which was significantly better than the best unsupervised reference method GBS (60.2\% mIoU). When compared to supervised methods, GrowSP-ForMS was comparable to PointNet (63.5\% mIoU), but far from RandLA-Net (74.1\% mIoU) and ResNet16FPN (78.5\% mIoU).

Finally, two ablation studies were performed. The first ablation study showed that our proposed improvements to GrowSP increased the test set mIoU of GrowSP-ForMS by 29.4 percentage points in comparison to the original model. Similarly, the second ablation study demonstrated that the addition of multiple reflectance channels can improve the accuracy of leaf--wood separation from ALS data. The full GrowSP-ForMS model that utilized all three reflectance channels for merging superpoints and clustering semantic primitives achieved a test set mIoU that was 5.6 perecentage points higher than the best performing monospectral model.

This study demonstrated the feasibility of utilizing unsupervised deep learning for leaf--wood separation from multispectral ALS data. Although the accuracy of GrowSP-ForMS is not at the level of modern supervised approaches, the fact that it performed comparably to an earlier fully supervised architecture and achieved state-of-the art performance in unsupervised leaf--wood separation is very promising, especially considering the various prospects for improving its performance in future work.

\section*{CRediT authorship contribution statement}

\textbf{Lassi Ruoppa:} Writing -- Original Draft, Conceptualization, Methodology, Software, Investigation, Data Curation, Visualization. \textbf{Oona Oinonen:} Writing -- Review \& Editing, Resources. \textbf{Josef Taher:} Writing -- Review \& Editing, Supervision, Resources. \textbf{Matti Lehtomäki:} Writing -- Review \& Editing, Supervision, Resources, Methodology, Data Curation. \textbf{Narges Takhtkeshha:} Writing -- Review \& Editing, Resources. \textbf{Antero Kukko:} Writing -- Review \& Editing, Investigation, Resources, Data Curation, Project administration, Funding acquisition. \textbf{Harri Kaartinen:} Writing -- Review \& Editing, Investigation, Resources. \textbf{Juha Hyyppä:} Writing -- Review \& Editing, Supervision, Resources, Project administration, Funding acquisition.

\section*{Acknowledgements}
We gratefully acknowledge the Research Council of Finland projects ``Forest-Human-Machine Interplay -- Building Resilience, Redefining Value Networks and Enabling Meaningful Experiences'' (decision no. 357908), ``Mapping of forest health, species and forest fire risks using Novel ICT Data and Approaches'' (decision no. 344755, including also CHIST-ERA-19-CES-001 funding), ``Collecting accurate individual tree information for harvester operation decision making'' (decision number 359554), ``Measuring Spatiotemporal Changes in Forest Ecosystem” (decision number 346382)'', ``High-performance computing allowing high-accuracy country-level individual tree carbon sink and biodiversity mapping'' (decision number 359203), ``Understanding Wood Density Variation Within and Between Trees Using Multispectral Point Cloud Technologies and X-ray microdensitometry'' (decision no. 331708), ``Capturing structural and functional diversity of trees and tree communities for supporting sustainable use of forests'', (decision no. 348644), ``Digital technologies, risk management solutions and tools for mitigating forest disturbances'' (decision no. 353264), the last two include also funding from the NextGenerationEU instrument, and the Ministry of Agriculture and Forestry grant ``Future Forest Information System at Individual Tree Level'' (VN/3482/2021).

\footnotesize{\bibliography{library}}

\clearpage
\normalsize
\begin{appendices}

\section{Optimization of reference models}

This appendix provides further details of the neural network training setups and hyperparameter optimization of the unsupervised reference methods.

\subsection{Neural network training details} \label{appendix:nn_details}

\subsubsection{GrowSP-ForMS}

Following the original GrowSP model, the number of semantic primitives and extracted features were set to $S=300$ and $K=128$ respectively. Based on initial experiments, the batch size was set to 16 and the weights given to reflectance features and geometric descriptors during semantic primitive clustering were $w_{\text{rgb}}=1$ and $w_{\text{geof}}=2$ respectively.

We trained GrowSP-ForMS for $E_{\text{pretrain}}=150$ epochs during the pretraining stage and $E_{\text{grow}}=60$ epochs during the growth stage, such that the number of superpoints decreased from $M^1=1500$ to $M^T=1200$. Although GrowSP was trained for a significantly higher number of epochs in the original article ($E_{\text{pretrain}}+E_{\text{grow}}\approx1000$), we observed that the training loss of GrowSP-ForMS stabilized quite early on our multispectral forest data set and increasing the number of training epochs did not improve the segmentation accuracy.

The number of superpoints in the initial merging $M^1=1500$ was proportional to that used by GrowSP ($M^1=80$) when accounting for the considerable difference in the number of initial superpoints $M^0$ between our forest environment data and indoor benchmark data sets such as S3DIS \citep{armeni20163d} and ScanNet \citep{dai2017scannet}. Although the original GrowSP training configuration decreased the number of superpoints by around 75\%, extensive testing revealed that such aggressive merging consistently reduced segmentation accuracy for forest data. As such, with GrowSP-ForMS we opted for a significantly shorter growth stage where the number of superpoints was only decreased by around 20\%.

\subsubsection{Supervised reference deep learning models}

As described in \autoref{section:training_nn}, an Adam \citep{kingma2014adam} optimizer with the cyclical learning rate scheduler of \cite{smith2017cyclical} was used for training the supervised baselines. Following \cite{kaijaluoto2022semantic}, we utilized the \emph{triangular2} scheduling policy, where the learning rate alternates linearly between a set base and a maximum value which is halved each time it is reached. The length of one cycle was set to 10 epochs and the base and maximum learning rates were determined with the learning rate test proposed by \cite{smith2017cyclical}. The batch sizes used for training each model were chosen based on initial experiments. The values of the training hyperparameters are listed in \autoref{table:training_params}, while other model specific data preprocessing steps and parameter values are described below.

\begin{table*}[!ht]
    \centering
    \caption{Hyperparameter values used for training the different supervised baselines. The learning rate values were determined using the learning rate test proposed by \cite{smith2017cyclical}. Note how the base and maximum learning rates of PointNet are the same, i.e. the learning rate is constant.} \bigskip
    \small
    \begin{tabular}{rcccc}
    \toprule
        \textbf{Model} & \textbf{Base learning rate} & \textbf{Maximum learning rate} & \textbf{Training epochs} & \textbf{Batch size} \\ \midrule \midrule
        PointNet \citep{qi2017pointnet} & 0.015 & 0.015 & 320 & 32 \\
        RandLA-Net \citep{hu2020randlanet} & 0.008 & 0.012 & 250 & 12 \\
        ResNet16FPN \citep{choy20194d} & 0.01 & 0.018 & 720 & 16 \\ \bottomrule
    \end{tabular}
    \label{table:training_params}
\end{table*}

\noindent \textbf{ResNet16FPN:} The data used for training ResNet16FPN was normalized as described in \autoref{section:data_preprocessing} and no modifications to the input data format were required. Following \cite{zhang2023growsp}, the voxel size used for discretizing the input point clouds was set to 0.05 m.

\noindent \textbf{RandLA-Net:} Since RandLA-Net requires point clouds of uniform size during training, we randomly sampled 80,000 points from each input cloud and only utilized the original clouds at prediction time. Additionally, we found that the input normalization strategy used by GrowSP was not adequate for training RandLA-Net as the loss did not converge. Consequently, following the semantic segmentation configuration of \cite{qi2017pointnet}, we applied standard min-max normalization to scale each feature to the range $[0,1]$. Finally, in accordance with the original article \citep{hu2020randlanet}, the number of neighbors used by the LocSE units was set to $k=16$.

\noindent \textbf{PointNet:} The training loss did not converge when using the cylindrical point clouds as input data for PointNet. As such, inspired by the semantic segmentation pipeline in the original PointNet article \citep{qi2017pointnet}, we further divided the cylinders into 2.8 m $\times$ 2.8 m blocks when training PointNet.

Similarly to RandLA-Net, PointNet requires uniformly sized inputs. As such, the blocks were subsampled to 10,000 points while training the model. Furthermore, the same min-max normalization strategy was applied for feature scaling.

\subsection{Hyperparameter optimization of unsupervised algorithms} \label{appendix:hyperparameter_optimization}

To optimize the hyperparameters of our unsupervised reference methods, LeWoS and GBS, we first determined a set of reasonable parameter values for both algorithms and subsequently utilized grid search to find the optimal hyperparameters. For each algorithm, the parameter combination that yielded the best training set mIoU was chosen.

LeWoS only requires one user-defined parameter, the feature difference threshold $N_{\text{z\_thres}}$. \cite{wang2020lewos} noted that the optimal value is generally within the range [0.1,0.2] and recommended the value 0.15. In order to explore the performance of LeWoS exhaustively, we defined the set of reasonable values as $N_{\text{z\_thres}}\in\{0.1,\ldots+0.025\ldots,0.4\}$, where $+d$ indicates several elements with a repeated difference of $d$.

Based on train split mIoU, the optimal parameter value for LeWoS was $N_{\text{z\_thres}}=0.2$ when inputs were entire forest scenes and $N_{\text{z\_thres}}=0.35$ when individual trees were provided as input.

In contrast to LeWoS, GBS has a total of four adjustable hyperparameters, the precursor direction difference threshold $\theta$, the interval length set $I$ and the linearity and cylindricity thresholds $L$ and $E$ \citep{tian2022graph}. For the interval length, we opted for the set $I=\{0.1, 0.2, 0.3, 0.5, 1\}$, since the authors suggested it for all trees with a height in the range [2 m, 60 m], which the trees in our data set fall within. Similarly, the authors recommend $\theta\in[0.2\pi,0.3\pi]$ when leaves are distributed at the ends of branches and $\theta\in[0.1\pi,0.2\pi]$ if the woody components are surrounded by leaves or leaf groups exhibit linear shapes. Consequently, we defined the set of reasonable parameter values as $\theta\in\{0.1\pi,0.2\pi,0.3\pi\}$. \cite{tian2022graph} empirically optimized the values of $L$ and $E$ to 0.9 and 0.2 respectively. In order to examine the effects of altering the values slightly, we defined the sets of reasonable values as $L\in\{0.85,0.9,0.95\}$ and $E\in\{0.15,0.2,0.25\}$.

Based on a grid search across the train split of our MS data set, the optimal GBS hyperparameter values were precursor direction difference threshold of $\theta=0.1\pi$, linearity threshold of $L=0.9$, and cylindricity threshold of $E=0.15$.

\section{Additional visualizations of segmentation results} \label{appendix:visual}

This appendix contains additional visual comparisons of the semantic segmentation results to support the results presented in \autoref{section:results}.

\subsection{Qualitative performance comparison on previously unseen unlabeled data}

\begin{figure*}[!hb]
\centering
\includegraphics[width=\textwidth]{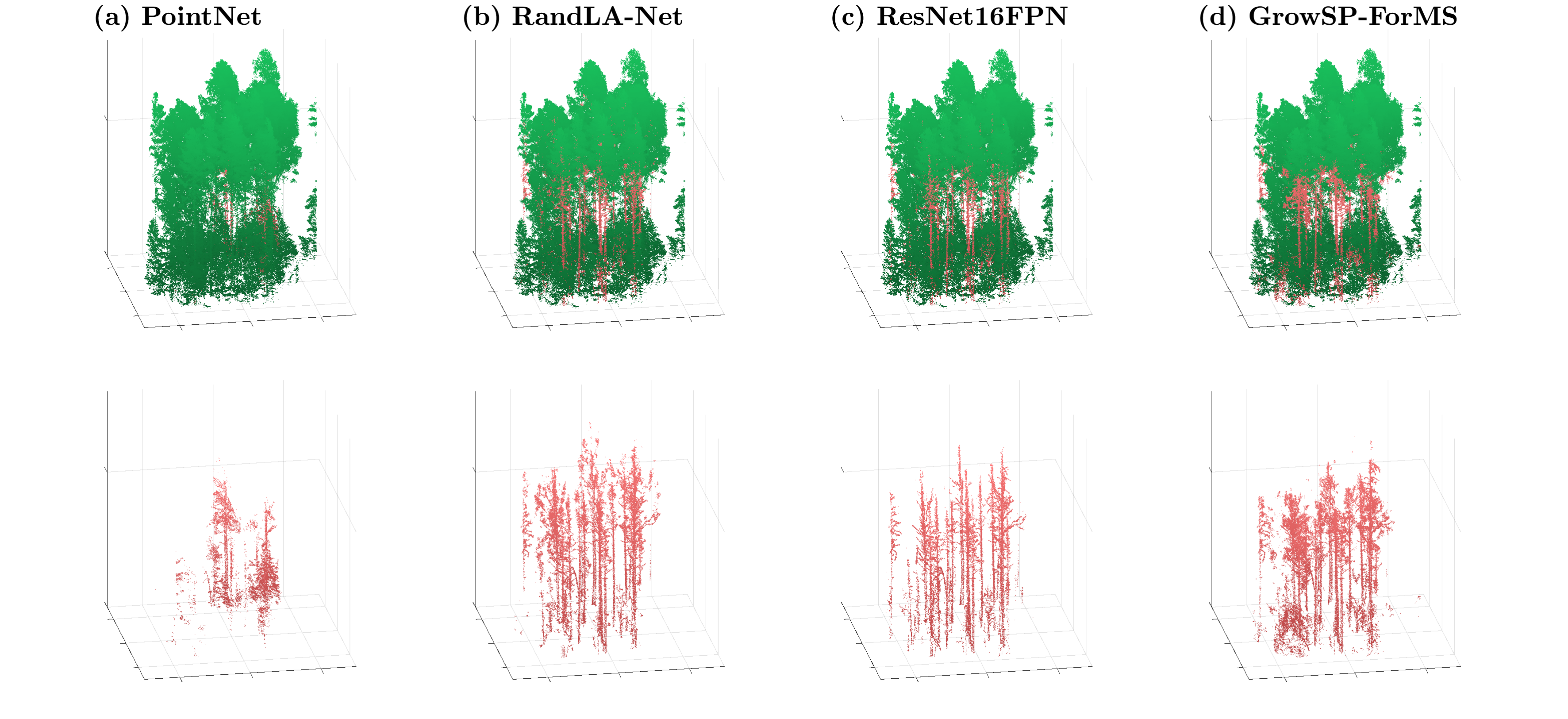}
\caption{Qualitative results of GrowSP-ForMS and our supervised reference models on a subsection of an unlabeled plot that was not used when training any of the models. The top row shows the full segmentation results and the row below contains only points classified as wood. Note the absence of ground truth data, which makes assessing the true quality of the predictions challenging.}
\label{fig:performance_comparison_2113}
\end{figure*}

\autoref{fig:performance_comparison_2113} displays a performance comparison between GrowSP-ForMS and our supervised reference leaf--wood separation models on a subsection of an unlabeled forest plot, which was not part of either the training or test set and was not utilized as auxiliary data when training GrowSP-ForMS. In other words, the input data is previously unseen for all models. Since no individual tree segments were available for the unlabeled data, the comparison was only performed for the supervised baselines, as a majority of the unsupervised reference algorithms require individual trees as input. It should be noted that the number of oversegmentation classes $|C_{\text{over}}|$ used by GrowSP-ForMS was set to 7 empirically since performance with the default value of 14 was insufficient.

As can be seen from the figure, the qualitative segmentation performance of RandLA-Net, ResNet16FPN, and GrowSP-ForMS remains similar to that achieved on the test set (see \autoref{fig:performance_comparison_all}), which suggests that all models generalize relatively well. As an exception, PointNet appears to perform notably worse on the unlabeled data, which is likely a result of the limited data set used for training. Interestingly, the limited size of the data set does not appear to have a notable effect on either RandLA-Net or ResNet16FPN. We note that although the segmentation results displayed in the figure appear to be of high quality, the lack of ground truth labels makes assessing the actual qualitative performance of each model relatively challenging.

\subsection{Qualitative performance comparison of ablated models} \label{appendix:ablated_visualization}

\begin{figure}[!hb]
\centering
\includegraphics[height=0.9\textheight]{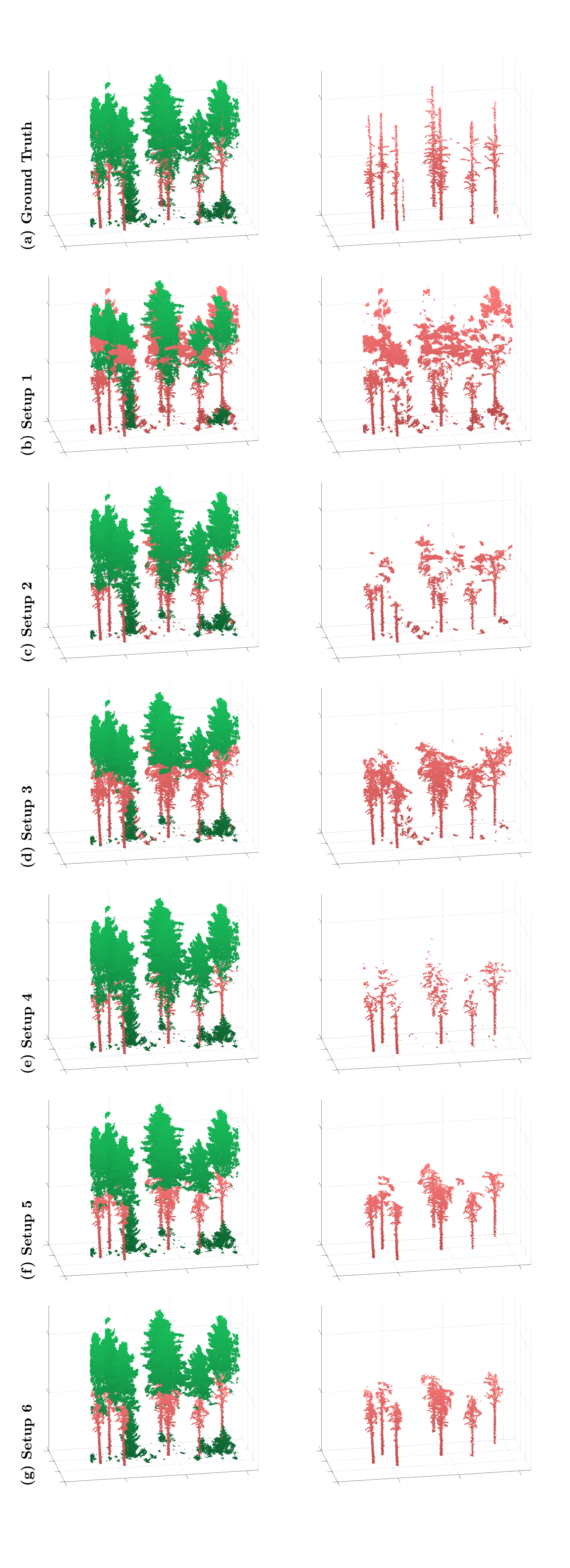}
\caption{Qualitative results of each ablated GrowSP-ForMS model on a subsection of the \textbf{test split} of plot \#1 in our multispectral data set. The first column the full segmentation results while the second column contains only points classified as wood. The different ablation setups are described in \autoref{table:ablation_params}.}
\label{fig:ablated_comparison}
\end{figure}

\autoref{fig:ablated_comparison} shows a visual comparison of the segmentation results from each ablated model from the first ablation study performed in \autoref{section:ablation_changes}, which evaluated the effects of our proposed improvements to the original GrowSP architecture. The improvements included in each ablated setup in the figure are detailed in \autoref{table:ablation_params}. The segmentation results shown in the figure suggest that each of our proposed changes to GrowSP also improves its qualitative performance, which is consistent with the quantitative performance metrics discussed in the ablation study.

\section{Additional quantitative performance metrics} \label{appendix:quantitative}

\subsection{Class specific segmentation accuracies}

\begin{figure}[!ht]
\centering
\includegraphics[height=0.9\textheight]{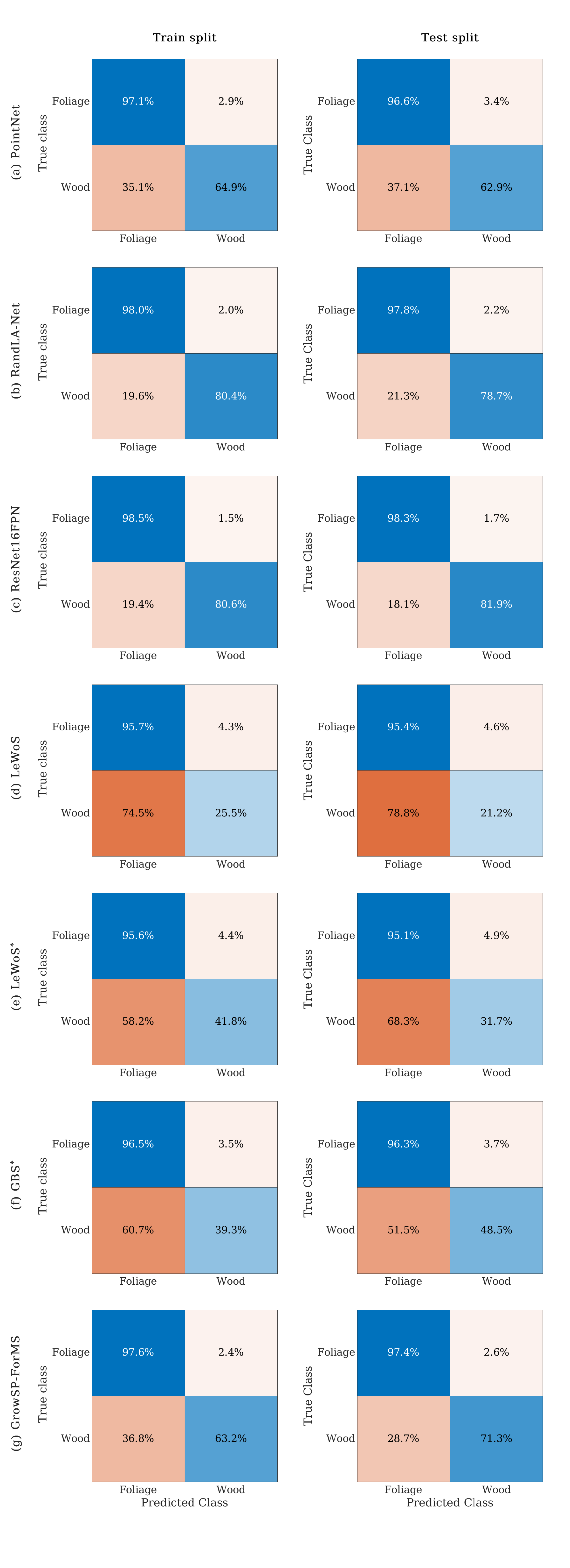}
\caption{Confusion matrices of GrowSP-ForMS and all reference leaf--wood separation models for both the train and test splits of our proposed multispectral data set. The asterisk ($^\ast$) denotes results when the inputs are individual trees, which are not strictly comparable to others due to the different data formats.}
\label{fig:confusion_matrix_all}
\end{figure}

\autoref{fig:confusion_matrix_all} lists the class-specific segmentation accuracies of the best-performing GrowSP-ForMS configuration and each reference leaf--wood separation method in the form of confusion matrices. Results are listed for both the train and test split of our multispectral data set.

\subsection{Training set accuracy metrics}

As a point of comparison to the test set performance metrics presented in \autoref{section:results}, this section lists the corresponding results on the train split of our multispectral data set. \autoref{table:model_comparison_train} lists quantitative performance metrics of all reference methods and the best-performing configuration of GrowSP-ForMS. Similarly, \autoref{table:ablation_scanners_train} shows the training set performance metrics of the ablated models from the second ablation study performed in \autoref{section:ablation_ms}, which assessed the benefits of using multispectral data for unsupervised leaf--wood separation.

\begin{table*}[!hb]
    \centering
    \caption{Quantitative results of different approaches on the \textbf{train split} of the proposed multispectral data set. The highest accuracy metrics in each category of supervision have been highlighted. The asterisk ($^\ast$) denotes results when the inputs are individual trees, which are not strictly comparable to others due to the different data formats.} \bigskip
    \small{
    \begin{tabular}{lccccc}
        \toprule
        \textbf{Model} & \textbf{OA (\%)} & \textbf{mAcc (\%)} & \textbf{mIoU (\%)} & \textbf{Foliage (\%)} & \textbf{Wood (\%)} \\ \midrule \midrule
        \textbf{Supervised methods} &&&&& \\ \midrule
        PointNet \citep{qi2017pointnet} & 96.0 & 81.0 & 65.5 & 96.0 & 35.0 \\
        RandLA-Net \citep{hu2020randlanet} & 97.3 & 89.2 & 75.2 & 97.3 & 53.1 \\
        ResNet16FPN \citep{choy20194d} & \textbf{97.7} & \textbf{89.5} & \textbf{79.0} & \textbf{97.6} & \textbf{60.4} \\ \midrule \midrule
        \textbf{Unsupervised methods} &&&&& \\ \midrule
        LeWoS \citep{wang2020lewos} & 93.8 & 60.6 & 51.8 & 93.8 & 9.8 \\
        LeWoS$^\ast$ \citep{wang2020lewos} & 94.6 & 68.7 & 53.4 & 94.6 & 12.1 \\
        GBS$^\ast$ \citep{tian2022graph} & 94.0 & 67.9 & 57.9 & 93.9 & 22.0 \\
        \textbf{GrowSP-ForMS (Ours)} & \textbf{96.1} & \textbf{80.4} & \textbf{68.1} & \textbf{96.0} & \textbf{40.2} \\ \bottomrule
    \end{tabular}}
    \label{table:model_comparison_train}
\end{table*}

\begin{table*}[!ht]
    \centering
    \caption{Ablation study on the effects of using mono- and multispectral data. The accuracies are reported for the \textbf{train split} of our multispectral data set. The best accuracy metrics have been highlighted.}
    \bigskip
    \footnotesize
    \begin{tabular}{cccccccc}
        \toprule
        \textbf{Scanner 1} & \textbf{Scanner 2} & \textbf{Scanner 3} & \textbf{oAcc (\%)} & \textbf{mAcc (\%)} & \textbf{mIoU (\%)} & \textbf{Foliage (\%)} & \textbf{Wood (\%)} \\ \midrule \midrule
        $\checkmark$ &&& 94.7 & 68.2 & 53.5 & 94.7 & 12.3 \\
        & $\checkmark$ && 94.7 & 71.8 & 62.1 & 94.6 & 29.5 \\
        && $\checkmark$ & 94.5 & 66.7 & 53.8 & 94.5 & 13.2 \\
        $\checkmark$ & $\checkmark$ && 92.5 & 66.0 & 59.8 & 92.3 & 27.3 \\
        $\checkmark$ && $\checkmark$ & 92.7 & 66.0 & 59.5 & 92.5 & 26.4 \\
        & $\checkmark$ & $\checkmark$ & 93.5 & 68.9 & \textbf{62.5} & 93.3 & \textbf{31.7} \\
        $\checkmark$ & $\checkmark$ & $\checkmark$ & \textbf{95.1} & \textbf{74.2} & 54.6 & \textbf{95.1} & 14.2 \\
        \bottomrule
    \end{tabular}
    \label{table:ablation_scanners_train}
\end{table*}

\end{appendices}

\end{document}